\documentclass{article}

\PassOptionsToPackage{numbers,compress}{natbib}
\usepackage[main,final]{neurips_2026}

\usepackage[utf8]{inputenc} 
\usepackage[T1]{fontenc}    
\usepackage{hyperref}       
\usepackage{url}            
\usepackage{booktabs}       
\usepackage{amsfonts}       
\usepackage{nicefrac}       
\usepackage{microtype}      
\usepackage{xcolor}         
\usepackage{multirow}
\usepackage{mathrsfs}
\usepackage{amsmath}
\usepackage[table]{xcolor}
\usepackage{arydshln}
\usepackage{graphicx}
\usepackage{subcaption}
\usepackage{wrapfig}
\usepackage{algorithm}
\usepackage{algpseudocode}
\usepackage{amssymb}
\usepackage{float}
\usepackage[most]{tcolorbox}
\usepackage{graphicx}
\definecolor{timabg}{RGB}{236,235,252}   
\definecolor{modelbg}{RGB}{232,232,232}  
\usepackage{makecell}

\newcommand{\sysname}{\textsc{Tima}}

\title{\sysname: Test-Time Internalization for Agentic Memory}

\author{%
    Xiaohang Sui$^{1}$, 
    Yongjian Fu\thanks{Corresponding Author.}\hspace{0.4em}$^{1}$, 
    Yizhe Zhao$^{1}$, 
    Sheng Yue$^{2}$, 
    Ju Ren$^{1,3}$ \\
    $^1$Department of Computer Science and Technology, Tsinghua University\\
    $^2$School of Cyber Science and Technology, Sun Yat-sen University\\
    $^3$State Key Laboratory of Internet Architecture\\
    {\tt\small \{suixh26, zhao-yz24\}@mails.tsinghua.edu.cn}, \\
    {\tt\small \{fuyongjian, renju\}@tsinghua.edu.cn}, \\
    {\tt\small yuesh5@mail.sysu.edu.cn}
}

\begin{document}

\maketitle

\begin{abstract}
Agentic memory is critical for accumulating and reusing experience in LLM agents across long interaction horizons and related tasks.
While latent-memory approaches offer a compact representation of such experience without incurring additional context overhead, existing methods are designed to encode transient in-context states or rely on pre-trained modules that remain fixed at test time to inject experiential representations. 
Consequently, an effective mechanism for continually refining reusable latent memory from execution experience is still missing in current agents.
To this end, we propose \sysname{}, a framework for test-time internalization that enables agents to acquire and update latent memory during interaction. 
\sysname{} couples a memory-augmented agent loop with an online-updated \textit{Internalizer} and an \emph{Internalizer Bank} for retrieval and continual refinement across episodes. 
Through self-supervised updates derived from execution trajectories, \sysname{} progressively encodes task-relevant solving patterns into compact latent memory, enabling both within-episode adaptation and cross-episode transfer.
Extensive experiments across eight benchmarks show that \sysname{} consistently outperforms strong baselines, surpassing A-Mem by up to 26.3\% and latent-memory baselines such as MemGen and Titans by up to 4.47\%--15.96\%.
Further analysis indicates that the learned latent memory remains stable under continual updates, generalizes well out of distribution, and exhibits interpretable clustering structure across \textit{Internalizers}.

\end{abstract}

\section{Introduction}
\begin{wrapfigure}{r}{0.34\columnwidth}
    \vspace{-9pt}
    \centering
    \includegraphics[width=\linewidth]{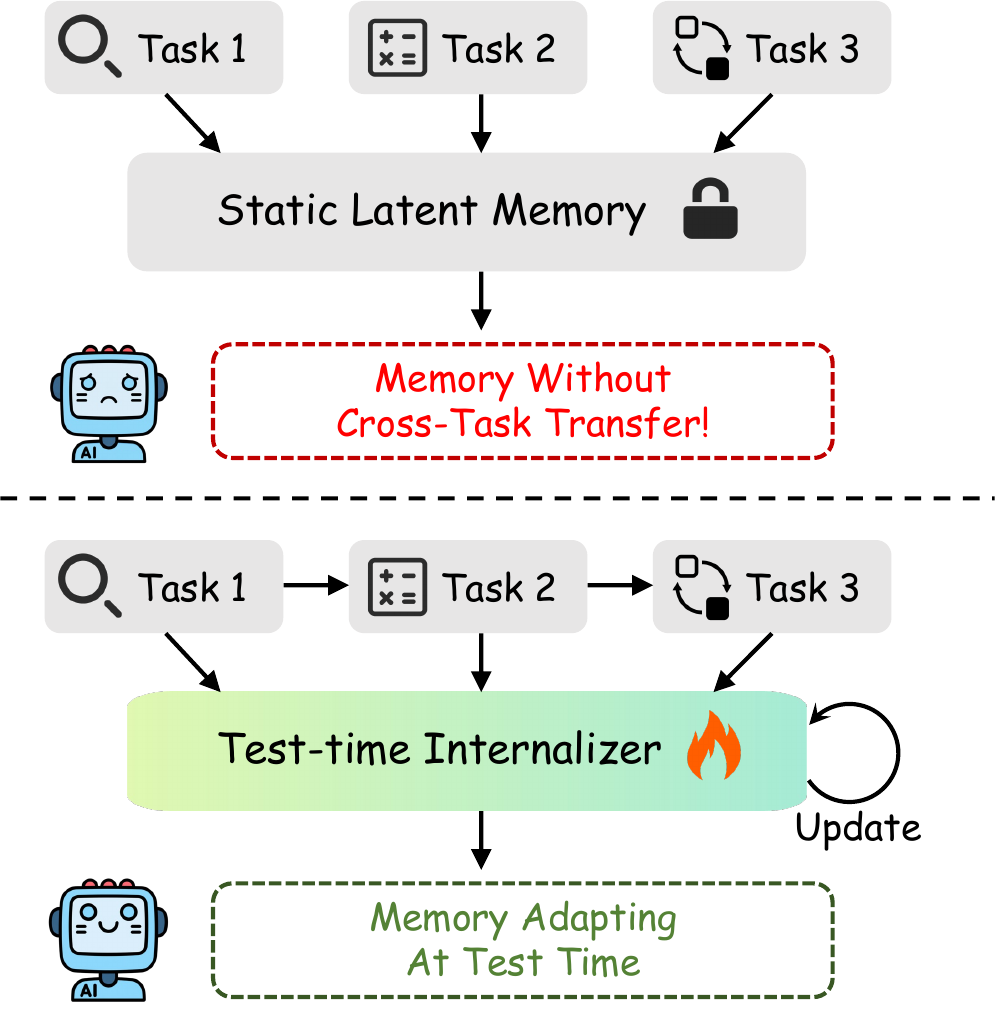}
    \vspace{-9pt}
    \caption{Overview of \sysname{}.}
    \label{fig:intro}
    \vspace{-10pt}
\end{wrapfigure}
Large Language Model (LLM)-based agents have recently demonstrated strong capabilities in complex problem solving and real-world interaction~\cite{yao2023react, shinn2023reflexion, jin2025searchr}.
However, many agentic tasks depend not only on reasoning over the current context, but also on the ability to maintain an informative \emph{memory}~\cite{hu2026memoryageaiagents} that retains and refines reusable experience across extended interactions and related tasks.
In practice, this remains a major challenge, as agent memory is typically bounded by a finite context window and lacks a persistent mechanism for compact experience accumulation~\cite{zhang2024surveymemorymechanismlarge, he2025humaninspiredperspectivessurveyai}.
Consequently, useful information from prior interactions is often not retained in a form that can be reliably transferred to subsequent tasks. 
These limitations highlight the need for memory mechanisms that support long-horizon reasoning, cross-task adaptation, and behavioral consistency in multi-turn environments.

Context-based methods~\cite{xu2026amem, kang-etal-2025-memoryos, chhikara2025mem0buildingproductionreadyai, agentkb, yu2026memagent, zhou2026mem1} store past interactions or condensed experience as natural-language text and reuse them by appending retrieved content to the input prompt. 
While intuitive and interpretable, such approaches consume the model's limited context budget~\cite{fang2026lightmem}, and their reliance on textual representations makes it difficult to compactly capture abstract task-solving patterns~\cite{wu2025autoscalingcontinuousmemorygui}.
Latent-memory methods instead encode information as continuous representations in the model's hidden space. 
However, existing latent memory systems~\cite{xu-etal-2025-softcot, tang2025razorattention, behrouz2024titanslearningmemorizetest, longmem, wang2024memoryllmselfupdatablelargelanguage} primarily model task-specific states and factual associations, and thus fall short of distilling reusable procedural solving patterns from accumulated agent trajectories.

Several recent works~\cite{fu2026latentmemcustomizinglatentmemory, wu2025towards, zhang2026memgen} attempt to address this gap by augmenting agents with learnable soft tokens during execution.
Although such methods move beyond explicit text-based memory, their memory modules remain fixed at test time, preventing newly generated solving experience from being incorporated into the latent memory itself. 
This reveals a more fundamental limitation of existing latent-memory methods: while previously encoded experience can be retrieved and reused, ongoing interaction experience cannot be transformed into updated internal representations.
Consequently, memory remains a static artifact of pre-training rather than a continually evolving substrate for adaptation during deployment.
A complementary line of research on test-time learning (TTL)~\cite{zuo2025ttrltesttimereinforcementlearning, sun2025learning, hu2025testtime} has shown promise in enabling models to adapt on the fly. 
However, existing TTL objectives are developed primarily for static language modeling~\cite{hubotter2025efficiently, behrouz2024titanslearningmemorizetest} and do not naturally extend to complex agentic workflows, where supervision arises from trajectories that interleave observations, reasoning traces, and actions rather than a homogeneous text stream.
Above observations pose a key challenge to us: 
\textit{How can an LLM agent acquire latent memory that is continuously refined during deployment through a self-supervised signal tailored to agentic workflows?}

To address these challenges, we propose \sysname{} (\textbf{T}est-time \textbf{I}nternalization of \textbf{M}emory for \textbf{A}gents), a framework that equips LLM agents with a lightweight, continuously updated latent memory evolved through test-time learning. 
At the core of \sysname{} is the \emph{Internalizer}, a lightweight latent memory module that produces a short sequence of memory tokens at each turn and is refined online through a self-supervised signal derived from agent execution. 
Its parameters are updated continuously at test time by minimizing the conditional negative log-likelihood of the agent's next composite action, yielding a learning signal grounded in the interleaved structure of the agent's own trajectories rather than in standard language modeling targets.
Unlike prior latent-memory modules that remain fixed during deployment, the \emph{Internalizer} continues to evolve with the live task stream and progressively internalizes reusable solving patterns, as shown in Figure~\ref{fig:intro}.
Furthermore, to efficiently organize accumulated experiences, \sysname{} maintains an \emph{Internalizer Bank}, which supports both retrieval of the most suitable \textit{Internalizer} and adaptive updating with new internalized knowledge. 

We evaluate \sysname{} on four task families covering eight benchmarks with multiple backbones, spanning mathematical reasoning, code generation, multi-hop question answering, and long-horizon interactive agents. 
Across these settings, \sysname{} consistently outperforms strong baselines, surpassing the textual-memory baseline A-Mem by up to 26.3\% on ALFWorld~\cite{shridhar2021alfworld} and latent-memory baselines such as MemGen and Titans by up to 4.47\%--15.96\%. 
Further analysis indicates that the learned latent memory remains stable under continual updates, generalizes well out of distribution, and exhibits interpretable clustering structure across \textit{Internalizers}. 
Ablations over key design choices further show that each component contributes consistent gains.

\section{Related Work}

\subsection{Memory Mechanisms for LLM Agents}
LLM agent memory has been studied along three complementary directions. 
Context-based methods store past interactions as natural-language memories and reuse them by appending retrieved content to the prompt~\cite{kang-etal-2025-memoryos, chhikara2025mem0buildingproductionreadyai}. 
Recent systems further organize memories into multi-level hierarchies~\cite{kang-etal-2025-memoryos, zhang2025memevolvemetaevolutionagentmemory} and improve retrieval via adaptive routing~\cite{yan2026adamemadaptiveusercentricmemory}, outcome-driven proxy reasoning~\cite{tan2026memsifteroffloadingllmmemory}, or event-centric indexing~\cite{hu2026memorymattersmoreeventcentric}. 
A second line internalizes experience into parameters, either by distilling successful trajectories from stronger models into smaller task-adaptive agents~\cite{chen2023fireactlanguageagentfinetuning, yin-etal-2024-agent} or by encoding experience in external parameterized modules~\cite{tack2024online, wise}.

Latent-memory approaches take a different route by encoding information as continuous representations in the model's hidden space. 
Early works~\cite{munkhdalai2024leavecontextbehindefficient, wang2024memoryllmselfupdatablelargelanguage} maintain compressive latent memories with self-updating mechanisms. 
In the agent setting, concurrent efforts~\cite{zhang2026memgen, fu2026latentmemcustomizinglatentmemory} extend the idea of soft reasoning~\cite{xu-etal-2025-softcot} to memory augmentation by injecting learnable latent token sequences during execution, thereby constructing continuous memory from the agent's current state. 
While these methods move beyond explicit text-based memory, they typically rely on externally pre-trained modules whose parameters remain frozen at deployment, limiting their ability to adapt to the evolving task distributions encountered at test time.

\subsection{Test-time Learning}
Test-time learning (TTL) adapts models at inference by updating parameters with supervision induced from test inputs. 
Tent~\cite{wang2021tent} introduced entropy minimization as a fully test-time adaptation objective, while later work developed more conservative variants to mitigate catastrophic forgetting~\cite{zhang2025come}. 
In language modeling, TTT layers~\cite{sun2025learning} replace attention with a learned hidden state updated by a self-supervised objective for each test sequence, and TLM~\cite{hu2025testtime} enables rapid domain adaptation by minimizing input perplexity.  
Titans~\cite{behrouz2024titanslearningmemorizetest} further extends this direction with a neural memory module driven by surprise metric.
However, these TTL objectives are primarily designed for static sequence modeling or single-turn reasoning. 
Agent trajectories interleave observations, reasoning traces, and actions, inducing dependencies that differ fundamentally from standard language modeling. 
This mismatch limits the direct applicability of existing TTL formulations to agent settings.
\sysname{} addresses this gap by equipping agents with latent memory that is continually refined through a test-time learning objective derived from their own execution trajectories.
\section{Preliminaries}
\textbf{Agentic Workflow.} 
We consider an LLM agent that engages in multi-turn interaction with an external environment to solve a given task.
At each turn $t$, the agent takes the task instruction $q$, all previous actions $a_{<t}$, and environmental observations $o_{<t}$, first produces a latent thinking process $z_t \sim \pi_\theta(\cdot \mid q, a_{<t}, o_{<t})$, and then executes an action $a_t \sim \pi_\theta(\cdot \mid q, a_{<t}, o_{<t}, z_t)$, where $\pi_\theta$ denotes the policy parameterized by LLM weights $\theta$. 
We denote $\hat{a}_t = (z_t, a_t)$ as the composite action at turn $t$. 
The environment then returns an observation $o_t \in \mathcal{O}$. 
This process repeats until a terminal condition is met, yielding an interaction trajectory $\tau = (q, \hat{a}_1, o_1, \hat{a}_2, o_2, \ldots, \hat{a}_T, o_T)$.

\textbf{Agent Memory System.} 
To augment LLM agents with persistent knowledge that supports long-horizon reasoning, memory-augmented agents~\cite{zhang2026memgen, xu2026amem, kang-etal-2025-memoryos, behrouz2024titanslearningmemorizetest} introduce an external memory $\mathcal{M}$ that persists across tasks. 
At each turn, the agent conditions on both past trajectory $\tau_{< t}$ and the memory information retrieved from $\mathcal{M}$ with $\tau_{< t}$:
\begin{equation}
    \hat{a}_{t} \sim \pi_\theta\bigl(\cdot \mid \tau_{< t}, \texttt{Retrieve}(\mathcal{M}, \tau_{< t})\bigr)
\end{equation}
Existing systems instantiate the memory module $\mathcal{M}$ in a variety of forms, including external textual knowledge stores that preserve explicit past interactions~\cite{chhikara2025mem0buildingproductionreadyai, kang-etal-2025-memoryos, xu2026amem}, and continuous memories that encode experience into parameters~\cite{wise, xia2026skillrlevolvingagentsrecursive} or latent states~\cite{wang2024memoryllmselfupdatablelargelanguage}, together with a corresponding retrieval function $\texttt{Retrieve}(\cdot)$ tailored to the underlying memory representation. 
Building on this view, we instantiate agent memory as a retrievable and continually evolving latent module that compactly internalizes task-solving patterns across interactions.
\section{Methodology}

\begin{figure}[t]
    \centering
    \includegraphics[width=0.96\linewidth]{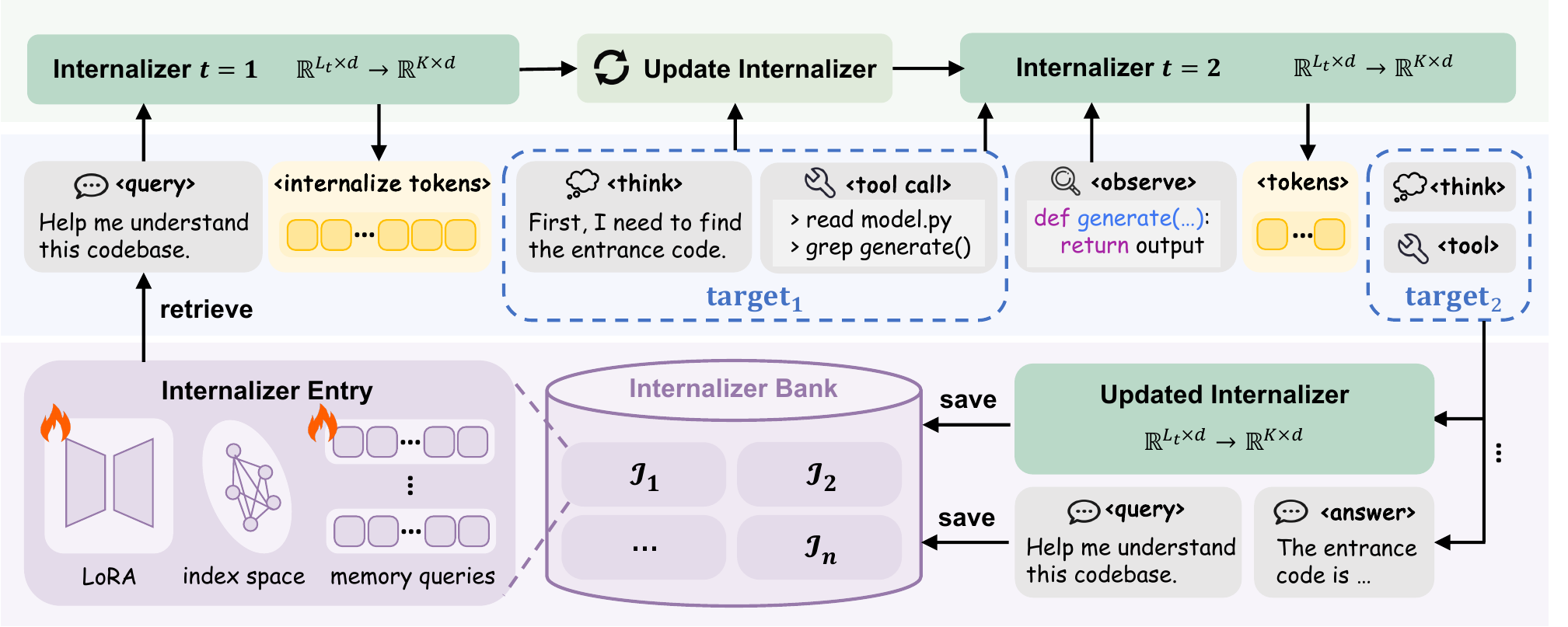}
    \caption{\textbf{The \sysname{} framework.} For each incoming task, \sysname{} retrieves an \textit{Internalizer} $\mathcal{I}$ from the \emph{Internalizer Bank} (bottom) and uses it to produce latent memory tokens that augment the agent's working loop (middle). After each turn, $\mathcal{I}$ is updated online through a test-time learning objective defined over the agent's own trajectory (top), and the updated parameters are written back to the bank for reuse on subsequent tasks.}
    \label{fig:framwork}
\end{figure}

We propose \sysname, a framework that enables LLM agents to acquire and refine reusable latent memory through test-time learning. 
As illustrated in Figure~\ref{fig:framwork}, \sysname{} comprises three tightly coupled components. 
First, it employs a memory-augmented agent loop, in which latent memory is injected at each interaction turn to inform the agent's subsequent reasoning and action (\S\ref{sec:agent_loop}). 
Second, it introduces an \textit{Internalizer}, a lightweight module that maps the agent's evolving interaction state to latent memory tokens and is updated online through a self-supervised test-time learning objective defined over the agent's self-generated trajectories (\S\ref{sec:internalizer}). 
Third, it maintains an \textit{Internalizer Bank}, which stores, retrieves, and updates task-relevant \textit{Internalizer}s across episodes, thereby enabling accumulated experience to persist and transfer beyond a single interaction (\S\ref{sec:bank}).

\subsection{Memory-Augmented Agent Loop}
\label{sec:agent_loop}

We consider an agent that interacts with the environment in a Think$\rightarrow$Act$\rightarrow$Observe loop~\cite{yao2023react}. 
At each turn $t$, upon receiving a new observation $o_t$ from the environment (or the initial task prompt $q$ at the start of the episode), 
\sysname{} invokes the \textit{Internalizer} to generate a latent memory segment that encodes compact experience relevant to the current interaction state. 
This memory segment is then incorporated into the current context, and the agent produces its next-step output conditioned on the resulting memory-augmented context.

Concretely, let $\mathbf{h}_{<t} = (\hat{a}_1, o_1, \ldots, \hat{a}_{t-1}, o_{t-1})$ denote the interaction history up to the $t$-th observation, where each composite action $\hat{a}_t = (z_t, a_t)$ consists of a reasoning trace $z_t$ and an executable action $a_t$.
Given the task query $q$ and a sliding window over the recent interaction history, the \textit{Internalizer} $\mathcal{I}_\phi$ produces $K$ latent memory tokens:
\begin{equation}
    m_t = \mathcal{I}_\phi\!\left(q, \mathbf{h}_{<t}[-w:]\right) \in \mathbb{R}^{K \times d}.
\end{equation}
where $d$ denotes the hidden dimension of the backbone LLM, and $\mathbf{h}_{<t}[-w:] = (\hat{a}_{t-w}, o_{t-w}, \ldots, \hat{a}_{t-1}, o_{t-1})$ denotes a sliding window over the recent interaction history, with $w$ being the window size.
These memory tokens are then incorporated into the model input, conditioning the prediction of the next composite action:
\begin{equation}
    \hat{a}_{t} \sim \pi_\theta\bigl(\cdot \mid q, \mathbf{h}_{<t}, m_t\bigr)
\end{equation}
The memory tokens $m_t$ encode internalized solving experience stored in the \textit{Internalizer} retrieved for the current query.

After the agent produces $\hat{a}_{t}$, \sysname{} updates the \textit{Internalizer} parameters $\phi$ using the resulting trajectory as a self-supervised learning signal. Through this test-time adaptation process, the \textit{Internalizer} progressively captures task-relevant solving patterns, allowing the latent memory to evolve alongside the agent's ongoing interaction.

\subsection{The \textit{Internalizer}}
\label{sec:internalizer}

\paragraph{Architecture.} 
The \textit{Internalizer} is a lightweight external module that encodes the agent's current interaction context into a fixed-length latent memory segment. 
Given the input token sequence, it produces $K$ continuous memory tokens in the same embedding space as the backbone LLM. 
The \textit{Internalizer} remains compact relative to the backbone, ensuring that both memory generation and test-time updates can be performed efficiently.

\paragraph{Test-Time Learning Objective.}
The key mechanism underlying \sysname{} is that the \textit{Internalizer} is updated online as interaction unfolds.
After each turn, the agent produces a composite action $\hat{a}_{t} = (z_t, a_t)$ conditioned on the memory-augmented context $(q, \mathbf{h}_{<t}[-w:], m_t)$.
We use the resulting reasoning-and-action trajectory as a self-supervised signal to update the \textit{Internalizer}$:$ intuitively, a more effective latent memory should render the agent's subsequent reasoning and action more predictable.
Formally, we minimize the conditional negative log-likelihood:
\begin{equation}
    \mathcal{L}^{(t)}_{\text{TTL}}(\phi) = -\log \pi_\theta\bigl(\hat{a}_{t} \mid q,\mathbf{h}_{<t}, m_t(\phi)\bigr)
    \label{eq:ttl}
\end{equation}
where $m_t(\phi)$ emphasizes the dependence of the memory tokens on the current \textit{Internalizer} parameters. 
The backbone parameters $\theta$ remain frozen, while gradients are applied only to the \textit{Internalizer} parameters $\phi$. 
This objective drives the \textit{Internalizer} to encode task-relevant solving patterns into its latent memory output, thereby improving the agent's ability to generate appropriate reasoning and actions for related tasks.

\paragraph{Entropy-Aware Objective Selection.}
Not all interaction turns provide equally reliable learning signals. 
Turns exhibiting high predictive uncertainty are more likely to induce noisy gradients, which may in turn compromise previously internalized experience.
Inspired by entropy-based sample filtering in TTA~\cite{niu2022efficient}, we therefore adopt a selection mechanism that prioritizes low-entropy turns during updating.

We compute the average token-level entropy of the agent's output over the think-and-act segment $\hat{a}_{t}$:
\begin{equation}
    E_t(\phi) = -\frac{1}{|\hat{a}_{t}|} \sum_{j=1}^{|\hat{a}_{t}|} \sum_{v \in \mathcal{V}} \pi_\theta(v \mid c_{<t},m_t(\phi),\hat{a}_{t}^{<j}) \log \pi_\theta(v \mid c_{<t},m_t(\phi),\hat{a}_{t}^{<j})
\end{equation}
where $\mathcal{V}$ is the vocabulary and $c_{<t}=(q,\mathbf{h}_{<t}[-w:])$. 
We define a turn-level weight:
\begin{equation}
    S_t = \exp\bigl(E_0 - E_t(\phi)\bigr) \cdot \mathbf{1}\bigl[E_t(\phi) < E_0\bigr]
    \label{eq:thresh}
\end{equation}
where $E_0$ is a determined entropy threshold. 
Turns with entropy above $E_0$ are discarded; those below contribute with weight proportional to their prediction confidence. 
The filtered objective is:
\begin{equation}
    \mathcal{L}_{\text{TTL}}^{\text{filtered}}(\phi) = \sum_{t=1}^T S_t \cdot \mathcal{L}^{(t)}_{\text{TTL}}(\phi)
\end{equation}

\paragraph{Update Procedure.}
We update the \textit{Internalizer} using a momentum-based rule that balances incorporating new experience against preserving prior knowledge, where the update takes the form:
\begin{equation}
\begin{aligned}
u &\leftarrow \mu u + (1-\mu)\,\nabla_{\phi}\mathcal{L}_{\text{TTL}}^{\text{filtered}}(\phi),\\
\phi &\leftarrow (1-\eta\omega)\,\phi - \eta\lambda\,u.
\end{aligned}
\label{eq:optim}
\end{equation}
where $u$ denotes the momentum-averaged update direction. $\mu$ controls momentum strength, with larger $\mu$ retaining more past update directions and making the \textit{Internalizer} evolve more conservatively. $\omega$ controls forgetting through parameter decay. $\eta$ and $\lambda$ jointly control the update magnitude.

\subsection{Internalizer Bank}
\label{sec:bank}

The \emph{Internalizer Bank} $\mathcal{B} = \{(\mathcal{Q}_i, \phi_i)\}_{i=1}^{N}$ maintains a collection of \textit{Internalizer} instances, each associated with a query set $\mathcal{Q}_i$ containing all task instructions it has previously served. 
The bank enables experience to persist and transfer across episodes.

\paragraph{Retrieval.}
When a new task query $q$ arrives, \sysname{} retrieves the most relevant \textit{Internalizer} by comparing the query $q$ with the entry key of each \textit{Internalizer} entry:
\begin{equation}
\mathbf{k}_i=\frac{1}{|\mathcal{Q}_i|}\sum_{q' \in \mathcal{Q}_i}\mathcal{E}(q'), \qquad
i^*=\arg\max_i \,\mathrm{sim}\!\left(\mathcal{E}(q), \mathbf{k}_i\right), \qquad
\tilde{\phi}\leftarrow\phi_{i^*}
\end{equation}
where $\mathcal{E}(\cdot)$ is the embedding encoder and $\text{sim}(\cdot, \cdot)$ denotes cosine similarity. 
The retrieval mechanism ensures $\tilde{\phi}$ carries internalized experience from prior tasks with a similar structure, providing the agent with a warm-started memory augmentation. 

\paragraph{Update and Initialization.}
After the agent completes the current task and the \textit{Internalizer} has been updated via TTL, the current query is registered into the associated query set: $\mathcal{Q}^* \leftarrow \mathcal{Q}^* \cup \{q\}$. 
The updated parameters $\tilde{\phi}$ replace the previous version in $\mathcal{B}$, ensuring the bank progressively accumulates solving experience.

When the bank is empty, the first task is handled by a freshly initialized \textit{Internalizer}, so memory emerges solely from the agent’s own test-time interaction. 
As the bank grows, new queries increasingly benefit from accumulated experience. 
To reduce cold start in practice, we initialize the bank with a small held-out task set, group tasks by query similarity, and use each cluster to briefly prime a dedicated \textit{Internalizer}. 
This grounds its memory tokens in representative trajectories and provides each entry with a coarse domain-specific prior before deployment on the live task stream.
\section{Experiments}

\subsection{Experimental Setup}
\label{sec:exp_setup}

\paragraph{Benchmarks.}
We evaluate \sysname{} across four task families.
For (i)~\textit{Mathematical Reasoning}, we use GSM8K~\cite{gsm8k} and MATH~\cite{hendrycks_math}. 
For (ii)~\textit{Code Generation}, we use KodCode~\cite{xu-etal-2025-kodcode} and BigCodeBench~\cite{bigcodebench}. 
For (iii)~\textit{Multi-Hop Question Answering}, we use HotpotQA~\cite{yang-etal-2018-hotpotqa} and 2WikiMultihopQA~\cite{ho-etal-2020-twowiki}, which require retrieval-conditioned reasoning over chains of supporting passages. 
And for (iv)~\textit{Multi-Turn Interactive Tasks}, we use ALFWorld~\cite{shridhar2021alfworld}, a text-based household simulator that tests long-horizon grounded action planning, and Webshop~\cite{webshop}, an online shopping simulator that tests instruction-conditioned navigation and purchase decisions under user specifications. 

\paragraph{Baselines.}
We compare \sysname{} against baselines spanning four categories. 
(i)~\textit{Vanilla agents} prompt the backbone without any external memory, including direct inference and a CoT-prompted variant~\cite{wei2022chain}. 
(ii)~\textit{Textual memory} inserts retrieved interaction history as natural language in the context, including RAG~\cite{rag} and A-Mem~\cite{xu2026amem}. 
(iii)~\textit{Latent Memory} methods include MemGen~\cite{zhang2026memgen} leveraging a tuned memory weaver and  the test-time surprise-updated Titans~\cite{behrouz2024titanslearningmemorizetest}. 
We additionally compare our method with Search-R1~\cite{jin2025searchr} as (iv)~\textit{RL-Augmented Memory} on multi-hop QA tasks. 

\paragraph{Implementation Details.}
We instantiate \sysname{} on two backbones of different scales, \textit{Qwen2.5-7B-Instruct}~\cite{qwen2025qwen25technicalreport} and \textit{Llama3.2-3B-Instruct}~\cite{touvron2023llamaopenefficientfoundation}, with identical decoding settings across all methods.
The \textit{Internalizer} is instantiated as Qwen2.5-1.5B equipped with LoRA adapters and a set of trainable memory query tokens, followed by a projection layer that aligns its outputs with the hidden space of the backbone model.
For the test-time learning objective we fix the number of memory tokens to $K=8$ and the sliding window size to $w=1$; the entropy threshold $E_0$ in Eq.~\ref{eq:thresh} is cold-started for the first 100 update steps on each task stream and is then set to the running $p_{50}$ quantile of the observed turn-level entropies. 
The \emph{Internalizer Bank} is seeded with 300 successful trajectories per benchmark, taken from AgentBank~\cite{song2024agentbankgeneralizedllmagents} for interactive tasks and from the official training split for static tasks, and strictly disjoint from the test set. 
We group them into 20 clusters by query similarity, and each cluster initializes one \texttt{seed} entry that provides a coarse domain prior and serves as a stable anchor that is never directly overwritten.
When a \texttt{seed} entry is retrieved for a new task, we copy it and update the copy through test-time learning on that task, which yields a \texttt{spec} entry that accumulates concrete test-time experience; a retrieved \texttt{spec} entry is updated in place.
Deployment thus begins with a coarse domain prior rather than a cold-start bank, and specialized experience grows without eroding the seeds.
Further details on hyperparameters and the full algorithm are deferred to Appendix~\ref{append:algo}.

\begin{table*}[t]
\centering
\small
\setlength{\tabcolsep}{6pt}
\caption{\textbf{Main results on reasoning benchmarks across two backbones.} Accuracy (\%) on math (GSM8K, MATH), code (KodCode, BigCodeBench), and multi-hop QA (2Wiki, HotpotQA). \textbf{Bold} denotes the best result per column; a dash (\textendash) indicates methods not evaluated on that benchmark. Search-R1 is a retrieval-specialized RL baseline that is applicable only to multi-hop QA.}
\begin{tabular}{lccccccc}
\toprule
\textbf{Method} & \multicolumn{2}{c}{\textbf{Math}} & \multicolumn{2}{c}{\textbf{Code}} & \multicolumn{2}{c}{\textbf{Multi-Hop QA}} & \textbf{Avg.} \\
\cmidrule(lr){2-3} \cmidrule(lr){4-5} \cmidrule(lr){6-7}
& GSM8K & MATH & KodCode & BigCodeBench & 2Wiki & HotpotQA & \\
\midrule
\multicolumn{8}{l}{\cellcolor{gray!20}\textbf{\textit{Qwen2.5-7B-Instruct}}} \\
Vanilla      & 90.33 & 72.60 & 52.39 & 36.99 & 21.61 & 16.74 & 48.44 \\
CoT          & 89.90 & 73.80 & 51.61 & 38.45 & 23.82 & 18.11 & 49.28 \\
\cmidrule(lr){1-8}
RAG          & 88.43 & 71.20 & 48.68 & 35.09 & 27.53 & 30.96 & 50.31 \\
A-Mem        & 89.73 & 76.40 & 51.02 & 39.47 & 39.02 & 33.69 & 54.89 \\
\cmidrule(lr){1-8}
Search-R1    & - & - & - & - & 40.06 & 41.84 & 40.95 \\
MemGen       & 91.60 & 79.40 & 65.92 & 40.64 & 39.61 & 38.56 & 59.29 \\
Titans       & 89.30 & 70.60 & 53.55 & 36.11 & 30.64 & 27.07 & 51.21 \\
\cmidrule(lr){1-8}
\rowcolor{blue!8}
\textsc{Tima}$_{w/o \, ttl}$ & 89.60 & 75.80 & 59.49 & 41.52 & 38.58 & 36.12 & 56.85 \\
\rowcolor{blue!8}
\textsc{Tima} & \textbf{93.47} & \textbf{80.40} & \textbf{66.31} & \textbf{43.27} & \textbf{41.04} & \textbf{43.03} & \textbf{61.25} \\

\midrule
\addlinespace[1pt]
\midrule

\multicolumn{8}{l}{\cellcolor{gray!20}\textbf{\textit{Llama3.2-3B-Instruct}}} \\
Vanilla      & 75.90 & 45.40 & 30.28 & 21.64 & 21.31 & 14.61 & 34.86 \\
CoT          & 77.00 & 47.80 & 29.89 & 23.25 & 19.14 & 15.48 & 35.43 \\
\cmidrule(lr){1-8}
RAG          & 73.97 & 42.80 & 29.40 & 19.74 & 23.83 & 26.68 & 36.07 \\
A-Mem        & 76.93 & 50.20 & 32.13 & 24.17 & 29.80 & 28.63 & 40.31 \\
\cmidrule(lr){1-8}
Search-R1    & - & - & - & - & 31.52 & 30.87 & 31.20 \\
MemGen       & 81.87 & 54.80 & \textbf{40.90} & 30.85 & 30.59 & 28.92 & 44.66 \\
Titans       & 78.93 & 42.20 & 31.16 & 23.68 & 26.59 & 24.25 & 37.80 \\
\cmidrule(lr){1-8}
\rowcolor{blue!8}
\textsc{Tima}$_{w/o \, ttl}$ & 79.27 & 49.60 & 35.73 & 31.29 & 29.45 & 27.85 & 42.20 \\
\rowcolor{blue!8}
\textsc{Tima} & \textbf{83.43} & \textbf{56.20} & 39.73 & \textbf{33.04} & \textbf{33.65} & \textbf{32.04} & \textbf{46.35} \\

\bottomrule
\end{tabular}
\label{tab:main_tab}
\end{table*}

\begin{table*}[t]
\centering
\small
\setlength{\tabcolsep}{6pt}
\caption{\textbf{Performance on ALFWorld and WebShop (Qwen2.5-7B).} For ALFWorld, we report success rate (\%) on the six sub-task types (\textit{pick}, \textit{look}, \textit{clean}, \textit{heat}, \textit{cool}, \textit{pick2}) and the overall average. For WebShop, we report both the average reward score and the average success rate (\%). \textbf{Bold} denotes the best result per column.}
\begin{tabular}{lccccccccc}
\toprule
\textbf{Method} & \multicolumn{7}{c}{\textbf{ALFWorld}} & \multicolumn{2}{c}{\textbf{WebShop}} \\
\cmidrule(lr){2-8} \cmidrule(lr){9-10}
& Pick & Look & Clean & Heat & Cool & Pick2 & Total & Score & Succ. \\
\midrule

\multicolumn{10}{l}{\cellcolor{gray!20}\textbf{\textit{Qwen2.5-7B-Instruct}}} \\

Vanilla & 11.9 & 5.2 & 10.3 & 19.6 & 4.9 & 16.1 & 10.9 & 26.1 & 7.6 \\
CoT     & 50.8 & 51.6 & 20.5 & 30.4 & 24.4 & 32.2 & 37.2 & 45.4 & 19.7  \\
\cmidrule(lr){1-10}
MemGen  & \textbf{93.2} & \textbf{69.0} & 74.4 & 69.6 & \textbf{85.4} & 58.1 & 76.3 & 70.6 & 54.9 \\
A-Mem   & 64.4 & 60.3 & 48.7 & 52.2 & 41.5 & 29.0 & 51.8 & 53.2 & 31.4 \\
\rowcolor{blue!8}
\sysname{} & 91.5 & 67.2 & \textbf{82.1} & \textbf{73.9} & \textbf{85.4} & \textbf{64.5} & \textbf{78.1} & \textbf{73.4} & \textbf{58.7} \\

\bottomrule
\end{tabular}
\label{tab:alf_web}
\end{table*}

\subsection{Main Results}

Table~\ref{tab:main_tab} presents the main results on six reasoning benchmarks across two backbones, Qwen2.5-7B-Instruct and Llama3.2-3B-Instruct, while Table~\ref{tab:alf_web} reports results on the ALFWorld interactive agent benchmark. 
We highlight four key observations from these results.

\textbf{(1) \sysname{} consistently outperforms all baselines across task families.} 
Table~\ref{tab:main_tab} shows that existing baselines exhibit clear trade-offs. 
Textual-memory approaches, including RAG and A-Mem, remain consistently limited across most benchmarks, indicating that explicit memory injection alone is insufficient for compactly capturing reusable solving patterns. 
Latent-memory baselines such as Titans and MemGen are markedly stronger overall, underscoring the effectiveness of latent memory representations for reasoning tasks. 
MemGen is the strongest among them and remains competitive across math, code, and multi-hop QA. 
However, this advantage is not uniformly sustained across settings. 
In contrast, \sysname{} consistently improves upon these baselines across nearly all benchmarks, with gains of up to 4.47\% over MemGen and 15.96\% over Titans.
On Qwen2.5-7B, it achieves the best result on all six benchmarks; on Llama3.2-3B, it ranks first on five of six and remains competitive on KodCode.
Relative to the vanilla backbone, \sysname{} improves the average accuracy by 12.81\% and 11.49\% on Qwen2.5-7B and Llama3.2-3B, respectively. 
These results indicate that online memory refinement through test-time internalization yields broader and more robust gains than either textual memory or fixed latent memory alone.

\textbf{(2) The margin over \sysname{} w/o TTL isolates the contribution of continual updating.}
Across both backbones, the full model consistently outperforms \sysname{} w/o TTL by 1.7\%--6.9\% on individual benchmarks, with an average improvement of roughly 4.3\%.
Since the two variants differ only in whether self-generated trajectories are used to perform test-time updates, this gap directly reflects the contribution of continual internalization.
In particular, warm-start retrieval from the \emph{Internalizer Bank} captures only part of the benefit, while online updating provides a further and consistent gain.

\textbf{(3) Offline-trained latent memory (MemGen) is the strongest non-TTL baseline but still trails.} 
MemGen substantially outperforms textual-memory approaches such as RAG and A-Mem. 
Nevertheless, \sysname{} outperforms MemGen on most benchmarks, with the largest gain appearing on HotpotQA under Qwen2.5-7B, where it improves accuracy by 4.47\%.
More broadly, \sysname{} shows its clearest advantages on benchmarks such as BigCodeBench, 2Wiki, and HotpotQA, suggesting that online latent-memory refinement is particularly beneficial in settings where task-solving patterns are less stable and must adapt to the test-time distribution.

\textbf{(4) \sysname{} transfers cleanly to long-horizon interactive agents.} 
Beyond single-query reasoning, Table~\ref{tab:alf_web} evaluates ALFWorld and WebShop, where each episode requires a grounded multi-step action sequence. 
On ALFWorld, \sysname{} achieves the best overall performance, improving over A-Mem by 26.3\% and over MemGen by 1.8\%. 
On WebShop, \sysname{} likewise attains the strongest results, exceeding MemGen 2.8 points in score and 3.8\% in success rate. 
The gains over MemGen on ALFWorld are concentrated on sub-tasks that require compositional procedural knowledge: \sysname{} improves \textit{clean}, \textit{heat}, and \textit{pick2} by 7.7\%, 4.3\%, and 6.4\%, respectively, while remaining near parity on simpler sub-tasks. 
Across both tables, the consistent gains support a unified view: self-supervised test-time updates from the agent's own trajectories strengthen the mapping from states to reasoning and action in a domain-adaptive manner, with the largest benefits arising where fixed offline memory cannot anticipate test-time solving patterns.

\subsection{Analysis}
\label{sec: analysis}

\paragraph{Online Learning Dynamics.}

\begin{figure}[t]
    \centering
     \begin{minipage}[t]{0.32\linewidth}
        \centering
        \includegraphics[width=\linewidth]{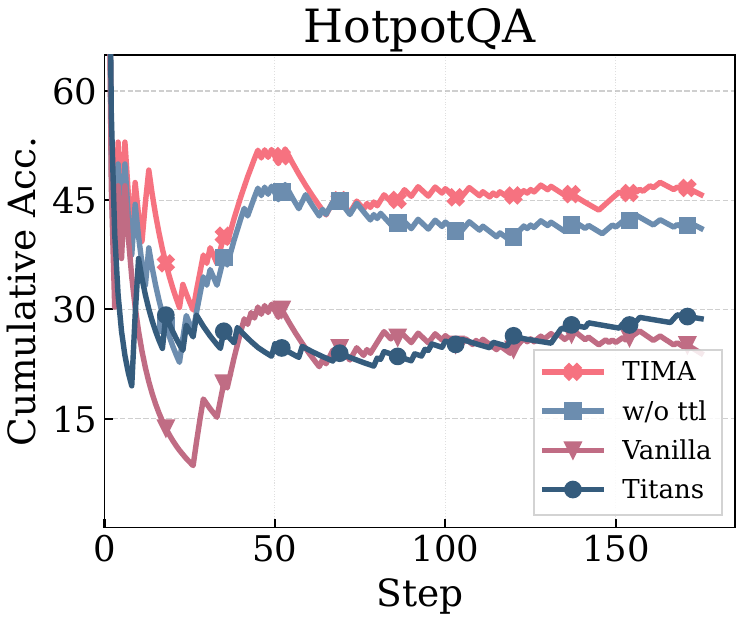}
    \end{minipage}
    \hfill
    \begin{minipage}[t]{0.32\linewidth}
        \centering
        \includegraphics[width=\linewidth]{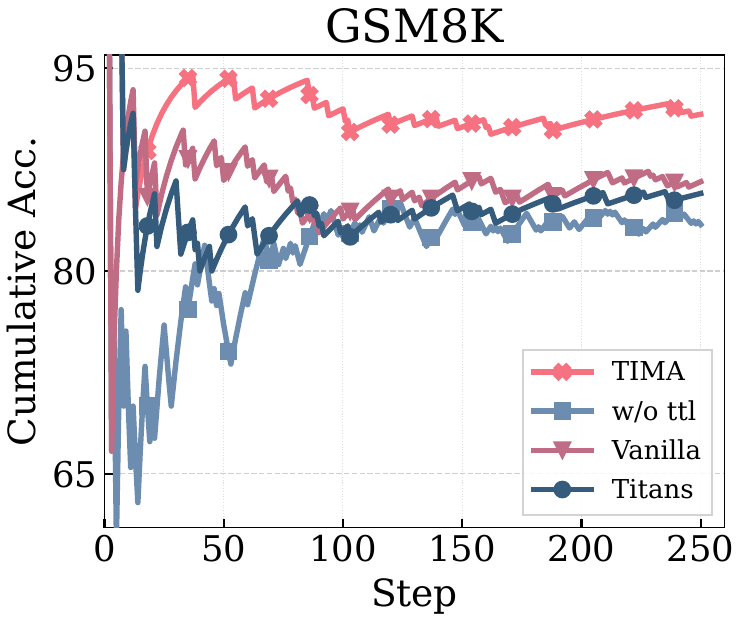}
    \end{minipage}
    \hfill
    \begin{minipage}[t]{0.32\linewidth}
        \centering
        \includegraphics[width=\linewidth]{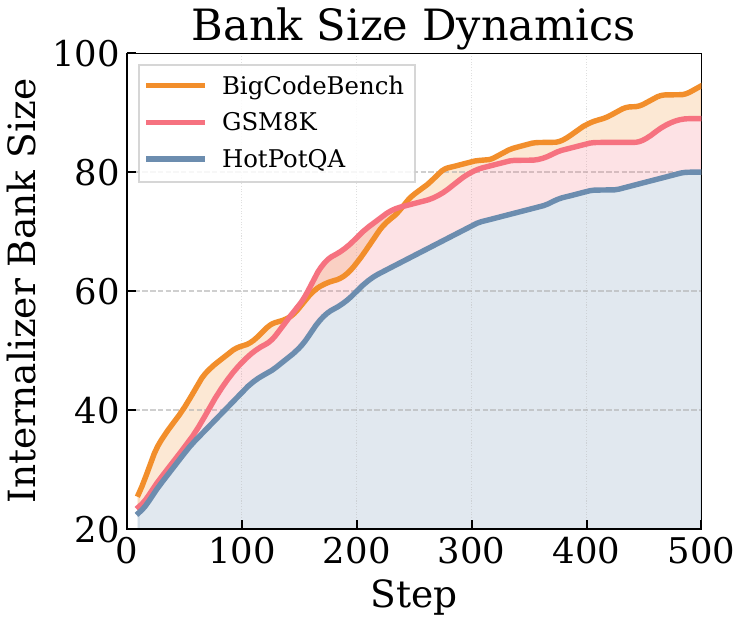}
    \end{minipage}
    \caption{\sysname{} maintains a persistent accuracy plateau while its \emph{Internalizer Bank} grows sub-linearly. \textbf{(Left, Middle)} Cumulative accuracy at each step on HotpotQA and GSM8K for \sysname{}, \sysname{} w/o TTL, Direct inference, and Titans. \textbf{(Right)} \emph{Internalizer Bank} size at each step on BigCodeBench, GSM8K, and HotpotQA.}
    \label{fig:dynamics}
\end{figure}

Figure~\ref{fig:dynamics} (left) tracks cumulative accuracy across the test stream on HotpotQA and GSM8K. 
\sysname{} reaches a higher accuracy plateau than all baselines and maintains a stable gap throughout the stream: after the initial warm-up, it stabilizes at roughly 45\% on HotpotQA, outperforming \sysname{} w/o TTL and all other baselines. 
Since \sysname{} and \sysname{} w/o TTL share the same pre-warmed \emph{Internalizer bank} and differ only in whether test-time updates are applied, this persistent gap is directly attributable to continual TTL rather than retrieval alone: online internalization provides gains beyond what a fixed bank can offer. 
In contrast, Titans remains near or below the vanilla baseline on both benchmarks despite updating its neural memory at test time, which we attribute to the mismatch between its token-level surprise objective and our next-action augmentation objective for agent success.
As shown in Figure~\ref{fig:dynamics} (right), the \emph{Internalizer Bank} grows sub-linearly and begins to saturate at later steps. 
As the stream progresses, newly arriving queries increasingly match existing \textit{Internalizer}s rather than spawning new ones, allowing the bank to specialize its entries in place instead of expanding indefinitely, a behavior that emerges naturally from test-time interaction without any explicit capacity cap.

\begin{figure}[t]
    \centering
    \includegraphics[width=0.96\linewidth]{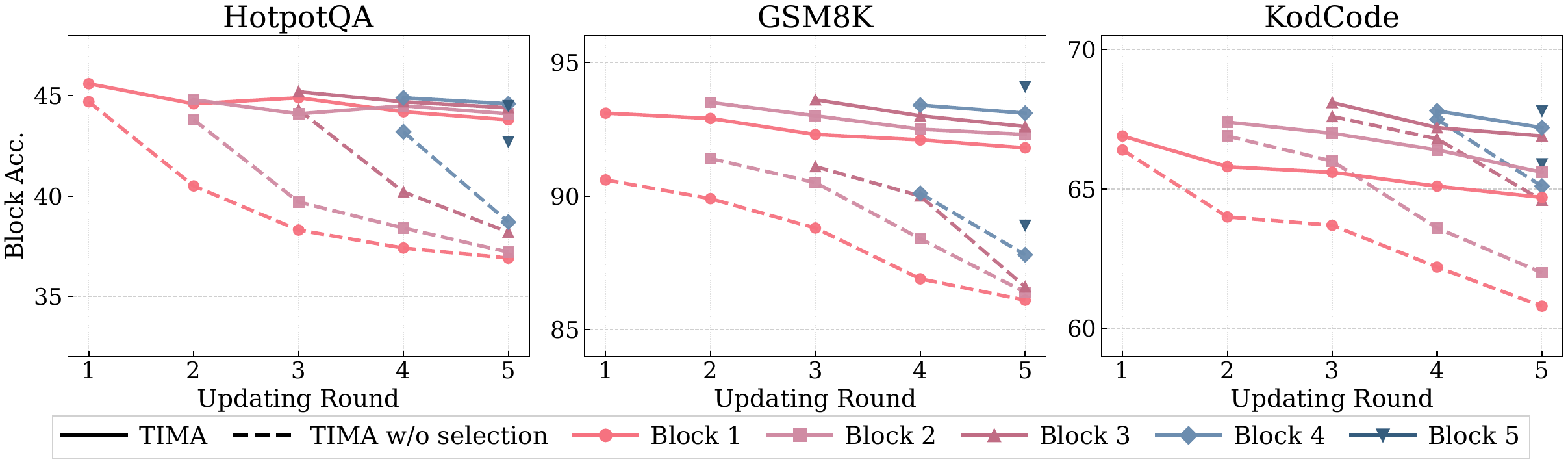}
    \caption{Block-wise accuracy on HotpotQA, GSM8K, and KodCode as the \textit{Internalizer} undergoes rounds of test-time updates on subsequent blocks. Each curve is one of five blocks; the marker on round $i$ is the block's first evaluation (evaluated immediately after TTL on that block), and later rounds evaluate the same block after TTL on subsequent blocks. Dashed lines use the same TTL objective without entropy filtering.}
    \label{fig:forgetting}
\end{figure}

\paragraph{Forgetting and Low-Entropy Selection Analysis.}
A central concern for any continuously updated memory is whether new experience overwrites prior knowledge. To probe this, we partition the test stream into five blocks and track block-level accuracy as the \textit{Internalizer} is updated over subsequent blocks, with and without the low-entropy selection objective. Figure~\ref{fig:forgetting} reports the resulting retention curves on multiple datasets. With low-entropy selection (solid lines), every block only performs minor accuracy drops (e.g., 45.6\% $\to$ 43.7\% on HotPotQA) over four additional update rounds; without it (dashed lines), the same blocks shows significant forgetting over the same horizon (e.g., 44.7\% $\to$ 36.9\% on HotPotQA), and the decay steepens as more updates are accumulated.

Both variants are evaluated from the same \sysname{} state before any cross-block update, isolating the gap as a direct effect of the update rule rather than retrieval or initialization. 
We attribute this separation to entropy acting as a noise filter: high-entropy trajectories provide ambiguous signals whose gradients accumulate destructively across rounds, while filtering them restricts updates to confident, reusable patterns aligned with the knowledge already stored in the \textit{Internalizer}. 
This retention holds across all three task families, suggesting that the mechanism generalizes beyond a single benchmark.

\begin{figure}[t]
    \centering
     \begin{minipage}[t]{0.49\linewidth}
        \centering
        \includegraphics[width=\linewidth]{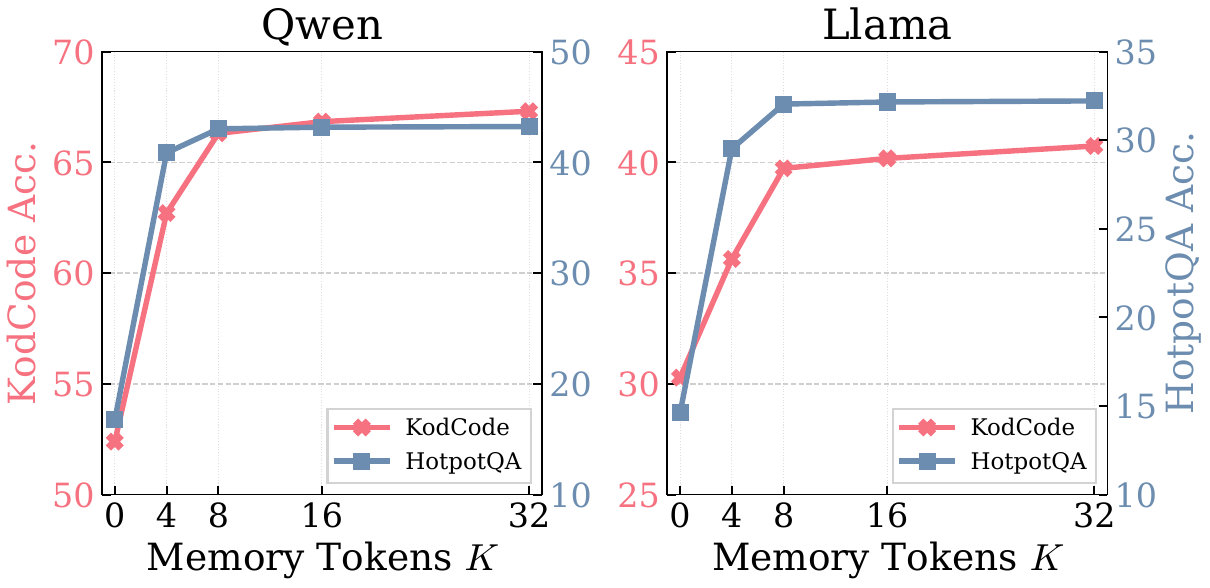}
    \end{minipage}
    \hfill
    \begin{minipage}[t]{0.49\linewidth}
        \centering
        \includegraphics[width=\linewidth]{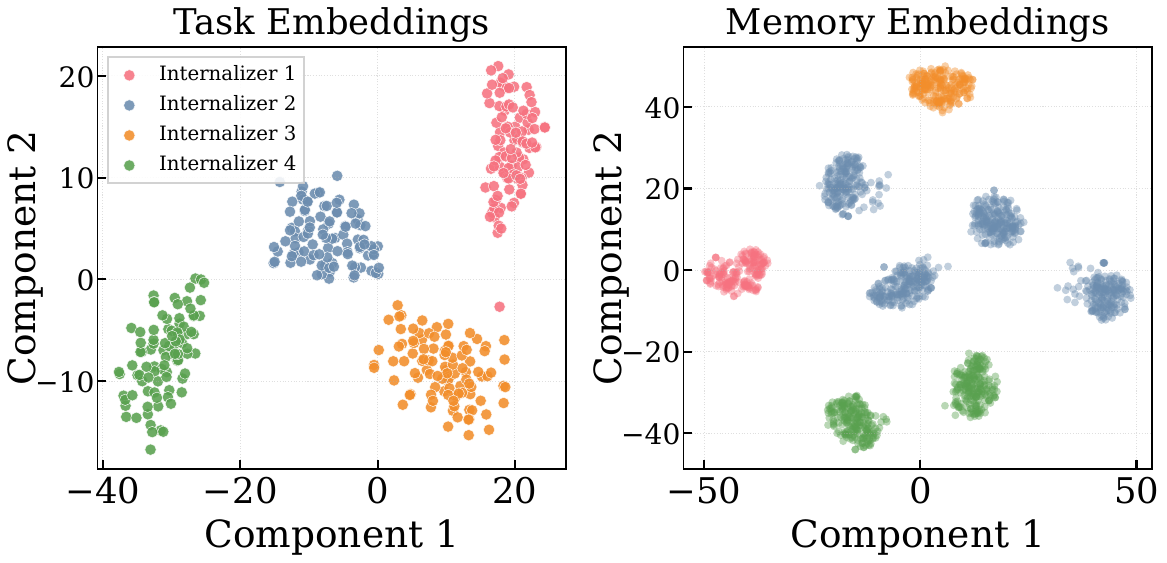}
    \end{minipage}
    \caption{\textbf{(Left) Memory token count ablation.} KodCode (red) and HotpotQA (blue) performance under $K \in \{0, 4, 8, 16, 32\}$ on \textit{Qwen2.5-7B-Instruct} and \textit{Llama3.2-3B-Instruct}. \textbf{(Right) Memory specialization on HotpotQA.} t-SNE projections of query embeddings and their memory-token outputs within on HotpotQA, colored by the \textit{Internalizer} each query is routed to.}
    \label{fig:abla_tsne}
\end{figure}

\paragraph{Memory Analysis.}

Figure~\ref{fig:abla_tsne} (right) contrasts the geometry of queries with that of their memory-token outputs. Queries of different types occupy well-separated regions in embedding space, and each region is handled by a distinct \textit{Internalizer}, so that after test-time updating, the bank has diversified across query types with each \textit{Internalizer} specialized to its own slice of the query distribution. The memory outputs on the right preserve this separation, and within some colors they break into additional sub-clusters, suggesting finer distinctions learned inside each \textit{Internalizer}. This embedding-level view matches the specialization-in-place behavior in Figure~\ref{fig:dynamics} that bank growth saturates once a small set of Internalizers covers the query-type structure of the stream, and further test-time updates refine each Internalizer internally rather than spawn new entries.

\paragraph{Ablation on Memory Tokens.}
Figure~\ref{fig:abla_tsne} (left) sweeps the number of memory tokens $K$ produced by the \textit{Internalizer}, from $K=0$ (no latent memory) up to $K=32$, on the two representative benchmarks KodCode and HotpotQA. Setting $K=0$ collapses \sysname{} to the Vanilla baseline on both backbones, attributing the Table~\ref{tab:main_tab} gains to the latent memory tokens. The first few tokens already deliver most of the benefit: moving from $K=0$ to $K=4$ raises HotpotQA accuracy by $24.1$ points on Qwen, suggesting that even a very compact memory captures the dominant solving-pattern signal. Beyond $K=8$, accuracy continues to rise monotonically as $K$ grows, indicating that \sysname{} keeps benefiting from additional memory capacity at diminishing marginal returns. Additional ablation studies are provided in Appendix~\ref{append:abla}.

\section{Conclusion}

We introduced \sysname{}, a test-time internalization framework that gives LLM agents a memory that adapts online to the current test stream. 
Our method maintains a small bank of \textit{Internalizers} that compress agent trajectories into latent memory tokens and are updated at test time with low-entropy selection. 
Across our benchmark suite and different backbone configurations, \sysname{} consistently outperforms textual-memory, offline-trained latent-memory, and RL-augmented baselines, while remaining stable under continual updates and generalizing across task families. 

\section*{Acknowledgement}


This research was supported in part by the National Natural Science Foundation of China under Grant No. 62432004, 62602337, the Fundamental and Interdisciplinary Disciplines Breakthrough Plan of the Ministry of Education of China under Grant No. JYB2025XDXM122, the Postdoctoral Innovative Talent Support Program under Grant No. BX2026290, and by a grant from the Guoqiang Institute, Tsinghua University.

{
\small
\bibliographystyle{plainnat}
\bibliography{references}
}

\clearpage

\appendix

\section{\sysname{} Algorithm}
\label{append:algo}
\begin{algorithm}[H]
\caption{\sysname{}: Test-Time Internalization for Agentic Memory}
\label{alg:tima}
\begin{algorithmic}[1]
\Require Task query $q$; frozen LLM $\pi_\theta$;
         Internalizer Bank $\mathcal{B} = \{(\mathcal{Q}_i, \phi_i, \texttt{type}_i)\}$
         where $\texttt{type}_i \in \{\texttt{seed}, \texttt{spec}\}$;
         entropy threshold $E_0$; window $w$; momentum $\mu$; update scale $\lambda$; decay $\omega$; lr $\eta$
\Ensure Final answer; updated bank $\mathcal{B}$

\Statex \textbf{// Phase 1: Retrieve Internalizer}
\State Compute $k_i = \frac{1}{|\mathcal{Q}_i|}\sum_{q'\in\mathcal{Q}_i}\mathcal{E}(q')$ for each entry; \quad $i^* = \arg\max_i\,\mathrm{sim}(\mathcal{E}(q), k_i)$
\State $\tilde{\phi} \leftarrow \phi_{i^*}$
\State Initialize momentum buffer $u \leftarrow 0$

\Statex \textbf{// Phase 2: Memory-Augmented Agent Loop}
\State Initialize history $\mathbf{h}_0 = \emptyset$
\For{turn $t = 1, 2, \ldots$ until terminal}
    \State $m_t = \mathcal{I}_{\tilde{\phi}}(q,\,\mathbf{h}_{<t}[-w:]) \in \mathbb{R}^{K\times d}$
    \State Sample $\hat{a}_{t} \sim \pi_\theta(\cdot \mid q,\mathbf{h}_{<t}, m_t)$
    \State Compute entropy $E_t(\tilde{\phi})$ over $\hat{a}_{t}$
    \If{$E_t < E_0$}
        \State $S_t = \exp(E_0 - E_t)$; \quad $\mathcal{L}_t = -S_t\cdot\log\pi_\theta(\hat{a}_{t}\mid q,\mathbf{h}_{<t}, m_t(\tilde{\phi}))$
        \State $u \leftarrow \mu u + (1-\mu)\,\nabla_{\tilde{\phi}}\mathcal{L}_t$
        \State $\tilde{\phi} \leftarrow (1-\eta\omega)\,\tilde{\phi} - \eta\lambda\, u$
    \EndIf
    \State Execute $a_{t}$; receive $o_{t}$; \quad $\mathbf{h}_{t} \leftarrow (\mathbf{h}_{<t}, \hat{a}_{t}, o_{t})$
\EndFor

\Statex \textbf{// Phase 3: Update Internalizer Bank}
\If{$\texttt{type}_{i^*} = \texttt{seed}$}
    \State Spawn new specialized entry: $\mathcal{B} \leftarrow \mathcal{B} \cup \bigl(\{q\},\; \tilde{\phi},\; \texttt{spec}\bigr)$
\Else \Comment{$\texttt{type}_{i^*} = \texttt{spec}$}
    \State $\mathcal{Q}_{i^*} \leftarrow \mathcal{Q}_{i^*} \cup \{q\}$; \quad recompute centroid $k_{i^*}$
    \State $\phi_{i^*} \leftarrow \tilde{\phi}$
\EndIf
\State \Return final answer; updated $\mathcal{B}$
\end{algorithmic}
\end{algorithm}

\section{More Experimental Details}
\label{append:details}

For the \textit{Internalizer} updating implementation. During each task, we collect the composite action training objectives at each turn, which are filtered by the entropy threshold and then used to update the \textit{Internalizer}. In practice, we implement the \textit{Internalizer} update with AdamW~\cite{loshchilov2019decoupledweightdecayregularization}, since its first-moment momentum and decoupled parameter decay align well with Eq.~\ref{eq:optim} and lead to more stable continual test-time updates. We adopt the corresponding trajectories in AgentBank~\cite{song2024agentbankgeneralizedllmagents} to seed our \emph{Internalizer Bank} from held-out samples for dynamics tasks while using the official training set for static tasks. Unless otherwise noted, reported numbers are averaged over multiple independent runs. Details of the hyperparameter setting are deferred to Table~\ref{tab:tima_hyperparams}.

The base parameters of Qwen2.5-1.5B are kept frozen. The context embeddings are first projected into the latent space of the \textit{Internalizer}, and $K$ trainable query embeddings are appended to them. After the sequence is processed by the \textit{Internalizer}, the hidden states at the query positions are extracted and mapped by an output projection into $K$ memory tokens in the hidden space of the backbone. The trainable parameters consist of the LoRA adapters, the input and output projections, and the query embeddings.

\begin{table}[H]
\centering
\small
\caption{Hyperparameters and implementation settings used in \sysname{}.}
\label{tab:tima_hyperparams}
\begin{tabular}{p{0.26\linewidth}p{0.66\linewidth}}
\toprule
\textbf{Settings} & \textbf{Hyperparameters} \\
\midrule

\textbf{Internalizer}
&
num\_latents $K$ = 8 \\
&
lora\_rank = 16 \\
&
lora\_alpha = 32 \\
&
lora\_target\_modules = [q\_proj, v\_proj] \\
&
lora\_dropout = None \\
&
latent\_init\_std = 0.02 \\
\midrule

\textbf{TTL Update}
&
max\_grad\_norm = 1.0 \\
&
optimizer\_learning\_rate $\eta$ = 1e-4\\
&
optimizer\_weight\_decay $\omega$ = 1e-3 \\
&
update\_strength $\lambda$ = 1.0 \\
&
momentum\_strength $\mu$ = 0.9 \\
&
entropy\_warm\_steps = 100 \\

\midrule

\textbf{Internalizer Bank}
&
retrieval\_encoder = sentence-transformers/all-MiniLM-L6-v2 \\
&
num\_seeds = 20 \\
&
seed\_samples = 300 \\
&
seed\_strategy = kmeans \\

\bottomrule
\end{tabular}
\end{table}

\section{Significance Analysis on Main Results}
\label{sec: significance}

\begin{table*}[t]
\centering
\small
\setlength{\tabcolsep}{4pt}
\caption{\textbf{Main results with standard deviation over five seeds.}
Accuracy (\%) on math, code, and multi-hop QA benchmarks.}
\begin{tabular}{lcccccc}
\toprule
\textbf{Method}
& \multicolumn{2}{c}{\textbf{Math}}
& \multicolumn{2}{c}{\textbf{Code}}
& \multicolumn{2}{c}{\textbf{Multi-Hop QA}} \\
\cmidrule(lr){2-3} \cmidrule(lr){4-5} \cmidrule(lr){6-7}
& GSM8K & MATH & KodCode & BigCodeBench & 2Wiki & HotpotQA \\
\midrule

\multicolumn{7}{l}{\cellcolor{gray!20}\textbf{\textit{Qwen2.5-7B-Instruct}}} \\
A-Mem
& 89.73{\scriptsize$\pm$0.67}
& 76.40{\scriptsize$\pm$0.71}
& 51.02{\scriptsize$\pm$0.76}
& 39.47{\scriptsize$\pm$0.95}
& 39.02{\scriptsize$\pm$1.23}
& 33.69{\scriptsize$\pm$1.35} \\
\cmidrule(lr){1-7}
Search-R1
& -- & -- & -- & --
& 40.06{\scriptsize$\pm$1.12}
& 41.84{\scriptsize$\pm$1.28} \\
MemGen
& 91.60{\scriptsize$\pm$0.68}
& 79.40{\scriptsize$\pm$0.67}
& 65.92{\scriptsize$\pm$0.55}
& 40.64{\scriptsize$\pm$0.81}
& 39.61{\scriptsize$\pm$0.98}
& 38.56{\scriptsize$\pm$0.90} \\
Titans
& 89.30{\scriptsize$\pm$1.25}
& 70.60{\scriptsize$\pm$1.18}
& 53.55{\scriptsize$\pm$1.08}
& 36.11{\scriptsize$\pm$1.36}
& 30.64{\scriptsize$\pm$1.72}
& 27.07{\scriptsize$\pm$1.53} \\
\cmidrule(lr){1-7}
\rowcolor{blue!8}
\textsc{Tima}
& \textbf{93.47}{\scriptsize$\pm$0.73}
& \textbf{80.40}{\scriptsize$\pm$0.78}
& \textbf{66.31}{\scriptsize$\pm$0.62}
& \textbf{43.27}{\scriptsize$\pm$1.04}
& \textbf{41.04}{\scriptsize$\pm$1.11}
& \textbf{43.03}{\scriptsize$\pm$0.88} \\

\midrule
\addlinespace[1pt]
\midrule

\multicolumn{7}{l}{\cellcolor{gray!20}\textbf{\textit{Llama3.2-3B-Instruct}}} \\
A-Mem
& 76.93{\scriptsize$\pm$0.72}
& 50.20{\scriptsize$\pm$0.84}
& 32.13{\scriptsize$\pm$1.10}
& 24.17{\scriptsize$\pm$1.09}
& 29.80{\scriptsize$\pm$1.22}
& 28.63{\scriptsize$\pm$1.31} \\
\cmidrule(lr){1-7}
Search-R1
& -- & -- & -- & --
& 31.52{\scriptsize$\pm$1.17}
& 30.87{\scriptsize$\pm$1.09} \\
MemGen
& 81.87{\scriptsize$\pm$0.67}
& 54.80{\scriptsize$\pm$0.65}
& \textbf{40.90}{\scriptsize$\pm$0.79}
& 30.85{\scriptsize$\pm$0.86}
& 30.59{\scriptsize$\pm$0.99}
& 28.92{\scriptsize$\pm$1.15} \\
Titans
& 78.93{\scriptsize$\pm$1.02}
& 42.20{\scriptsize$\pm$1.16}
& 31.16{\scriptsize$\pm$1.41}
& 23.68{\scriptsize$\pm$1.32}
& 26.59{\scriptsize$\pm$1.47}
& 24.25{\scriptsize$\pm$1.78} \\
\cmidrule(lr){1-7}
\rowcolor{blue!8}
\textsc{Tima}
& \textbf{83.43}{\scriptsize$\pm$0.82}
& \textbf{56.20}{\scriptsize$\pm$0.81}
& 39.73{\scriptsize$\pm$1.03}
& \textbf{33.04}{\scriptsize$\pm$1.12}
& \textbf{33.65}{\scriptsize$\pm$1.15}
& \textbf{32.04}{\scriptsize$\pm$1.20} \\
\bottomrule
\end{tabular}
\label{tab:main_results_std}
\end{table*}

Table~\ref{tab:main_results_std} reports the means and standard deviations behind the main results over five seeds. \sysname{} is stable across runs: its standard deviation lies between 0.62 and 1.11 on Qwen2.5-7B-Instruct and between 0.81 and 1.20 on Llama3.2-3B-Instruct, comparable to MemGen (0.55--0.98 and 0.65--1.15) and consistently lower than Titans (1.08--1.72 and 1.02--1.78), the other method that updates its memory at test time. The continual \emph{Internalizer} updates therefore do not add run-to-run instability. Relative to MemGen, \sysname{} gains between $+0.39$ (KodCode) and $+4.47$ (HotpotQA) points on Qwen2.5-7B-Instruct, and between $+1.40$ and $+3.12$ points on five of the six benchmarks on Llama3.2-3B-Instruct, with the largest margin on HotpotQA for both backbones. The exception is KodCode on Llama3.2-3B-Instruct, where MemGen is ahead by $1.17$ points. Relative to Search-R1, \sysname{} improves 2Wiki and HotpotQA by $+0.98$ and $+1.19$ points on Qwen2.5-7B-Instruct and by $+2.13$ and $+1.17$ points on Llama3.2-3B-Instruct, without any retrieval-specific RL training.

\begin{table*}[t]
\centering
\small
\setlength{\tabcolsep}{4pt}
\caption{\textbf{Results on interactive benchmarks with standard deviation.} ALFWorld success rate (\%), WebShop score, and WebShop success rate (\%) on Qwen2.5-7B-Instruct and \textbf{Llama3.2-3B-Instruct}.}
\begin{tabular}{lccccc}
\toprule
\textbf{Metric} & \textbf{Vanilla} & \textbf{CoT} & \textbf{MemGen} & \textbf{A-Mem} & \textsc{Tima} \\
\midrule

\multicolumn{6}{l}{\cellcolor{gray!20}\textit{\textbf{Qwen2.5-7B-Instruct}}} \\
ALFWorld
& 10.9{\scriptsize$\pm$1.1}
& 37.2{\scriptsize$\pm$1.5}
& 76.3{\scriptsize$\pm$1.6}
& 51.8{\scriptsize$\pm$2.0}
& 78.1{\scriptsize$\pm$1.7} \\
WebShop Score
& 26.1{\scriptsize$\pm$1.5}
& 45.4{\scriptsize$\pm$2.3}
& 70.6{\scriptsize$\pm$2.0}
& 53.2{\scriptsize$\pm$2.1}
& 73.4{\scriptsize$\pm$1.9} \\
WebShop Succ.
& 7.6{\scriptsize$\pm$1.0}
& 19.7{\scriptsize$\pm$1.7}
& 54.9{\scriptsize$\pm$2.1}
& 31.4{\scriptsize$\pm$1.9}
& 58.7{\scriptsize$\pm$1.6} \\

\midrule
\addlinespace[1pt]
\midrule

\multicolumn{6}{l}{\cellcolor{gray!20}\textit{\textbf{Llama3.2-3B-Instruct}}} \\
ALFWorld
& 8.4{\scriptsize$\pm$1.2}
& 28.7{\scriptsize$\pm$1.8}
& 57.9{\scriptsize$\pm$1.5}
& 38.8{\scriptsize$\pm$2.3}
& 59.6{\scriptsize$\pm$1.6} \\
WebShop Score
& 12.1{\scriptsize$\pm$1.4}
& 33.9{\scriptsize$\pm$1.6}
& 58.4{\scriptsize$\pm$1.9}
& 45.1{\scriptsize$\pm$2.4}
& 62.7{\scriptsize$\pm$1.7} \\
WebShop Succ.
& 4.2{\scriptsize$\pm$1.2}
& 16.1{\scriptsize$\pm$1.9}
& 42.3{\scriptsize$\pm$2.1}
& 25.7{\scriptsize$\pm$2.0}
& 48.6{\scriptsize$\pm$2.2} \\
\bottomrule
\end{tabular}
\label{tab:interactive_results_std}
\end{table*}

Table~\ref{tab:interactive_results_std} gives the same breakdown for the interactive benchmarks. The standard deviations are larger than on static tasks (1.0--2.4 points across all methods), since multi-turn rollouts compound the stochasticity of both the policy and the environment, yet the ordering of methods is unchanged: \sysname{} is the best method on every metric for both backbones. On Qwen2.5-7B-Instruct, it improves over MemGen by $+1.8$ on ALFWorld, $+2.8$ on WebShop score, and $+3.8$ on WebShop success rate; on Llama3.2-3B-Instruct the corresponding gains are $+1.7$, $+4.3$, and $+6.3$. The largest gains appear on WebShop success rate, and the $6.3$-point margin on Llama3.2-3B-Instruct is about three times the standard deviation of either method. The gap to A-Mem is considerably larger, reaching $+26.3$ on ALFWorld and $+27.3$ on WebShop success rate with Qwen2.5-7B-Instruct.


\begin{table*}[t]
\centering
\caption{\textbf{One-sided paired $t$-tests of \textsc{Tima} against MemGen and Search-R1 over five seeds.} Positive $t$-values indicate that \textsc{Tima} outperforms the baseline; $p<0.05$ denotes statistical significance.}
\label{tab:ttest}
\setlength{\tabcolsep}{4pt}
\resizebox{\textwidth}{!}{%
\begin{tabular}{l cccccccc cc}
\toprule
 & \multicolumn{8}{c}{\textbf{vs.\ MemGen}} & \multicolumn{2}{c}{\textbf{vs.\ Search-R1}} \\
\cmidrule(lr){2-9} \cmidrule(lr){10-11}
\textbf{Metric} & \textbf{GSM8K} & \textbf{MATH} & \textbf{KodCode} & \textbf{BigCodeBench} & \textbf{2Wiki} & \textbf{HotpotQA} & \textbf{ALFWorld} & \textbf{WebShop} & \textbf{2Wiki} & \textbf{HotpotQA} \\
\midrule
\multicolumn{11}{>{\columncolor{modelbg}}l}{\textbf{\textit{Qwen2.5-7B-Instruct}}} \\
\textbf{$t$-value} & 3.880 & 2.341 & 2.434 & 7.288 & 2.675 & 12.164 & 2.359 & 5.990 & 2.486 & 2.555 \\
\textbf{$p$-value} & 0.009 & 0.040 & 0.036 & 0.001 & 0.028 & $<$0.001 & 0.039 & 0.002 & 0.034 & 0.031 \\
\midrule
\multicolumn{11}{>{\columncolor{modelbg}}l}{\textbf{\textit{Llama3.2-3B-Instruct}}} \\
\textbf{$t$-value} & 3.268 & 3.264 & $-$2.612 & 2.982 & 6.667 & 4.349 & 2.236 & 5.423 & 2.536 & 3.173 \\
\textbf{$p$-value} & 0.015 & 0.015 & 0.971 & 0.021 & 0.001 & 0.006 & 0.044 & 0.003 & 0.032 & 0.017 \\
\bottomrule
\end{tabular}%
}
\end{table*}

Table~\ref{tab:ttest} complements these statistics with one-sided paired $t$-tests over the five seeds ($\mathrm{df}=4$). Against MemGen and Search-R1, \sysname{} is significantly better ($p<0.05$) in all ten comparisons on Qwen2.5-7B-Instruct and in nine of ten on Llama3.2-3B-Instruct, with the largest $p$-value among the significant results being $0.044$. The evidence is strongest on HotpotQA ($t=12.164$, $p<0.001$) and BigCodeBench ($t=7.288$, $p=0.001$) with Qwen2.5-7B-Instruct, and on 2Wiki ($t=6.667$, $p=0.001$) and WebShop ($t=5.423$, $p=0.003$) with Llama3.2-3B-Instruct. The comparisons against Search-R1 on 2Wiki and HotpotQA are significant on both backbones. The only comparison in which \sysname{} is not significantly better is KodCode on Llama3.2-3B-Instruct, where MemGen has a $1.17$-point higher mean ($t=-2.612$).

\section{Experience Transferring Analysis   }
\label{append:ood}
We acquire the \emph{Internalizer Bank} through our framework on one source dataset and evaluate directly on held-out targets without further adaptation, using Qwen2.5-7B-Instruct for all methods. Two source--target groups are considered: KodCode$\to$\{BigCodeBench, HumanEval\} for code generation, and HotpotQA$\to$\{2Wiki, GPQA\} for multi-hop reasoning. 

\begin{figure}[t]
\centering
\includegraphics[width=\linewidth]{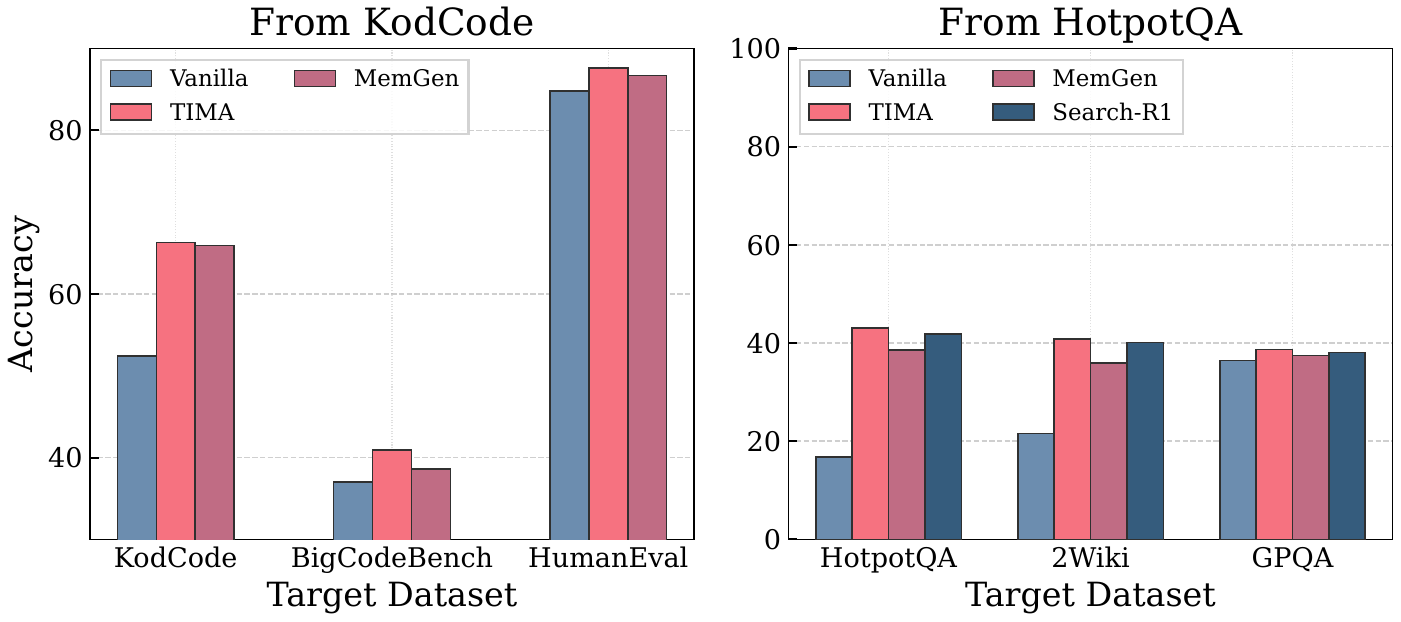}
\caption{\textbf{Experience Transferring Performance.} For \sysname{}, the \emph{Internalizer Bank} is accumulated on KodCode (left) or HotpotQA (right) and evaluated on held-out targets. \sysname{} retains its lead across various sets, which underscores the generalization ability of our method. }
\label{fig:ood}
\end{figure}

As shown in Figure~\ref{fig:ood}, \sysname{} leads on both in-domain sources and preserves its edge under moderate shift. Relative to MemGen, the strongest latent-memory baseline, \sysname{} improves by $+2.3$ points when transferring from KodCode to BigCodeBench and by $+4.9$ points when transferring from HotpotQA to 2Wiki, indicating that the internalized memory retains a recoverable prior across adjacent task distributions rather than overfitting to a single dataset. A few additional observations are also worth highlighting. First, \sysname{} stays above Vanilla on every one of the six target settings, so keeping the bank active never incurs a measurable transfer cost even when the target drifts far from the source distribution. Second, on the QA side, \sysname{} remains ahead of Search-R1 across all three targets without any retrieval-specific RL training, suggesting that test-time internalization can match the cross-distribution robustness of a retrieval-specialized policy at a substantially lower training cost.

\section{Ablation Study}
\label{append:abla}
\subsection{Ablation on LoRA Rank}

\begin{table}[t]
\centering
\caption{\textbf{Effect of LoRA rank $r$ in the \textit{Internalizer}.} Performance on GSM8K, KodCode, and 2Wiki with the \textit{Qwen2.5-7B-Instruct} backbone. Bold marks the rank used in the main experiments.}
\label{tab:lora_rank}
\begin{tabular}{lccc}
\toprule
\textbf{LoRA Rank} & \textbf{GSM8K} & \textbf{KodCode} & \textbf{2Wiki} \\
\midrule
$r=4$  & 91.63 & 62.84 & 38.02 \\
$r=6$  & 92.07 & 63.58 & 38.74 \\
$r=8$  & 92.57 & 64.92 & 39.55 \\
\rowcolor{gray!12}
$r=16$ \; \textbf{(Default)} & 93.47 & 66.31 & 41.04 \\
$r=24$ & \underline{93.60} & 66.52 & \underline{41.26} \\
$r=32$ & 93.57 & \underline{66.73} & 41.19 \\
\bottomrule
\end{tabular}
\end{table}

Table~\ref{tab:lora_rank} ablates the LoRA rank $r$ used inside the \textit{Internalizer} on three representative datasets (GSM8K, KodCode, 2Wiki), keeping all other hyperparameters fixed at their main-experiment values. Accuracy rises steadily from $r=4$ to $r=16$ on every dataset, indicating that the \textit{Internalizer} needs a minimum amount of low-rank capacity to absorb the reusable solving patterns carried by each trajectory. The sensitivity to rank is not uniform across tasks: KodCode and 2Wiki benefit noticeably more from a larger rank than GSM8K does, suggesting that code generation and multi-hop question answering depend more heavily on latent adaptation capacity. Beyond $r=16$, the curves flatten, and the higher ranks $r=24$ and $r=32$ yield only minor, task-specific fluctuations; although the per-column best (underlined) is attained at $r=24$ or $r=32$, the gap to $r=16$ is minor on every dataset. We therefore adopt $r=16$ as the default setting in the main experiments, trading a negligible accuracy margin for a substantially lower test-time update cost.


\subsection{Ablation on Low-Entropy Threshold}

\begin{table}[t]
\centering
\caption{\textbf{Effect of the entropy threshold $E_0$ used by low-entropy selection.} $E_0$ is set to a percentile of the running turn-level entropy distribution. }
 \label{tab:entropy}
  \begin{tabular}{lccc}
    \toprule
    $E_0$ \textbf{Setting}  & \textbf{GSM8K} & \textbf{HotpotQA} & \textbf{KodCode} \\
    \midrule
    $p_{10}$                         & 91.87 & 39.24 & 63.29 \\
    $p_{25}$                       & 92.83 & 41.38 & 65.04 \\
    \rowcolor{gray!12}
    $p_{50}$ \; \textbf{(Default)}  & \textbf{93.47} & \textbf{43.03} & \textbf{66.31} \\
    $p_{75}$                       & 92.77 & 40.70 & 64.65 \\
    No filtering                 & 91.57 & 37.10 & 62.51 \\
    \bottomrule
  \end{tabular}
\end{table}

Table~\ref{tab:entropy} sweeps the entropy threshold $E_0$ used by the low-entropy selection rule in Eq~\ref{eq:thresh}, expressed as the percentile of the running turn-level entropy distribution at which the threshold is placed. $E_0 = p_{50}$, which keeps the bottom half of the turns for TTL, is the best setting on all three benchmarks, showing that filtering out roughly half of the highest-entropy turns strikes the right balance between the updating signal and noise turns. The two directions away from this peak are not symmetric: raising the threshold to $p_{75}$ hurts more than lowering it to $p_{25}$, indicating that the destructive effect of admitting additional noisy high-entropy turns outweighs the cost of keeping fewer updates. Disabling filtering altogether produces the weakest result on every benchmark, and the resulting HotpotQA accuracy aligns with the w/o selection setting in Figure~\ref{fig:forgetting}. 

\subsection{Ablation on Auxiliary Parameters}

\begin{table}[t]
\centering
\caption{\textbf{Backbone-only online LoRA TTL ablation.} To isolate the effect of online updating from the additional 1.5B auxiliary parameters, \emph{TTL on Backbone} applies online LoRA test-time learning directly to the backbone with the same objective, while removing the memory-token injection mechanism. Best results are in bold.}
\label{tab:backbone_ttl}
\begin{tabular}{lcccc}
\toprule
\textbf{Method} & \textbf{GSM8K} & \textbf{KodCode} & \textbf{HotpotQA} & \textbf{ALFWorld} \\
\midrule
Vanilla         & 90.33 & 52.39 & 16.74 & 10.9 \\
TTL on Backbone & 89.67 & 55.11 & 24.83 & 50.6 \\
\textsc{Tima}   & \textbf{93.47} & \textbf{66.31} & \textbf{43.03} & \textbf{78.1} \\
\bottomrule
\end{tabular}
\end{table}

Table~\ref{tab:backbone_ttl} separates the effect of online updating from that of the additional 1.5B auxiliary parameters. Applying the same low-entropy TTL objective directly to the backbone with online LoRA, without memory-token injection, improves KodCode by $2.72$ points, HotpotQA from 16.74 to 24.83, and ALFWorld from 10.9 to 50.6, while GSM8K slightly decreases from 90.33 to 89.67. Test-time learning is therefore useful on its own for multi-turn and knowledge-intensive tasks, but it does not account for the full gain: \sysname{} outperforms the backbone-only variant by $+3.80$ on GSM8K, $+11.20$ on KodCode, $+18.20$ on HotpotQA, and $+27.5$ on ALFWorld. The drop on GSM8K is consistent with online updates altering the pretrained weights that carry the model's reasoning ability, whereas \sysname{} keeps the backbone frozen and confines adaptation to the \emph{Internalizer}, which supplies memory through latent tokens. The improvement thus comes from combining memory-token injection with test-time internalization rather than from extra parameters or online updating alone.

\subsection{Ablation on Internalizer Size}

\begin{table}[t]
\centering
\caption{\textbf{Internalizer-size ablation on Llama3.2-3B-Instruct.} We replace the default Qwen2.5-1.5B-Instruct Internalizer with the smaller Qwen2.5-0.5B-Instruct, which introduces fewer additional parameters and lower computational overhead. Best results are in bold.}
\label{tab:internalizer_size}
\begin{tabular}{lccc}
\toprule
\textbf{Internalizer} & \textbf{GSM8K} & \textbf{KodCode} & \textbf{HotpotQA} \\
\midrule
\rowcolor{gray!12}
Qwen2.5-1.5B \textbf{(Default)} & \textbf{83.43} & \textbf{39.73} & \textbf{32.04} \\
Qwen2.5-0.5B           & 81.30 & 38.66 & 28.24 \\
\bottomrule
\end{tabular}
\end{table}

Table~\ref{tab:internalizer_size} replaces the default Qwen2.5-1.5B-Instruct \emph{Internalizer} with the smaller Qwen2.5-0.5B-Instruct on Llama3.2-3B-Instruct. Reducing the \emph{Internalizer} to one third of its default size lowers accuracy by $2.13$, $1.07$, and $3.80$ points on GSM8K, KodCode, and HotpotQA, respectively, indicating that the capacity of the \emph{Internalizer} translates into the quality of the generated latent memory. The drop is largest on HotpotQA, suggesting that multi-hop reasoning relies more on the expressiveness of the memory generator than arithmetic or code generation does. The 0.5B variant introduces fewer additional parameters and lower computational overhead, and it still remains ahead of Titans on all three benchmarks (78.93, 31.16, and 24.25), so it is a practical option when latency is the main concern. We use the 1.5B \emph{Internalizer} by default for the best accuracy.

\subsection{Ablation on Context-window Size}

\begin{table}[t]
\centering
\caption{\textbf{Effect of the context-window size $w$.} $w$ denotes the number of most recent action--observation pairs used as context for the Internalizer. Best results are in bold.}
\label{tab:context_window}
\begin{tabular}{lccc}
\toprule
\textbf{Context window size ($w$)} & \textbf{GSM8K} & \textbf{KodCode} & \textbf{2Wiki} \\
\midrule
\rowcolor{gray!12}
1 \textbf{(Default)} & \textbf{93.47} & \textbf{66.31} & \textbf{41.04} \\
2           & 93.10 & 65.53 & 40.70 \\
4           & 92.47 & 65.14 & 40.11 \\
\bottomrule
\end{tabular}
\end{table}

Table~\ref{tab:context_window} varies the number of most recent action--observation pairs $w$ provided to the \emph{Internalizer}. Accuracy decreases monotonically as the window grows on all three benchmarks: moving from $w=1$ to $w=4$ lowers GSM8K, KodCode, and 2Wiki by $1.00$, $1.17$, and $0.93$ points, respectively, with $w=2$ falling in between. The \emph{Internalizer} thus draws the most useful signal from the latest interaction step, while earlier pairs lengthen the input and dilute the current step with older context. The changes are small in magnitude, so \sysname{} is not sensitive to this hyperparameter. Since $w=1$ is both the most accurate and the cheapest setting, we adopt it as the default.

\section{More Experimental Results}

\subsection{Forgetting Analysis w/o TTL}

\begin{figure}[t]
\centering
\includegraphics[width=0.94\linewidth]{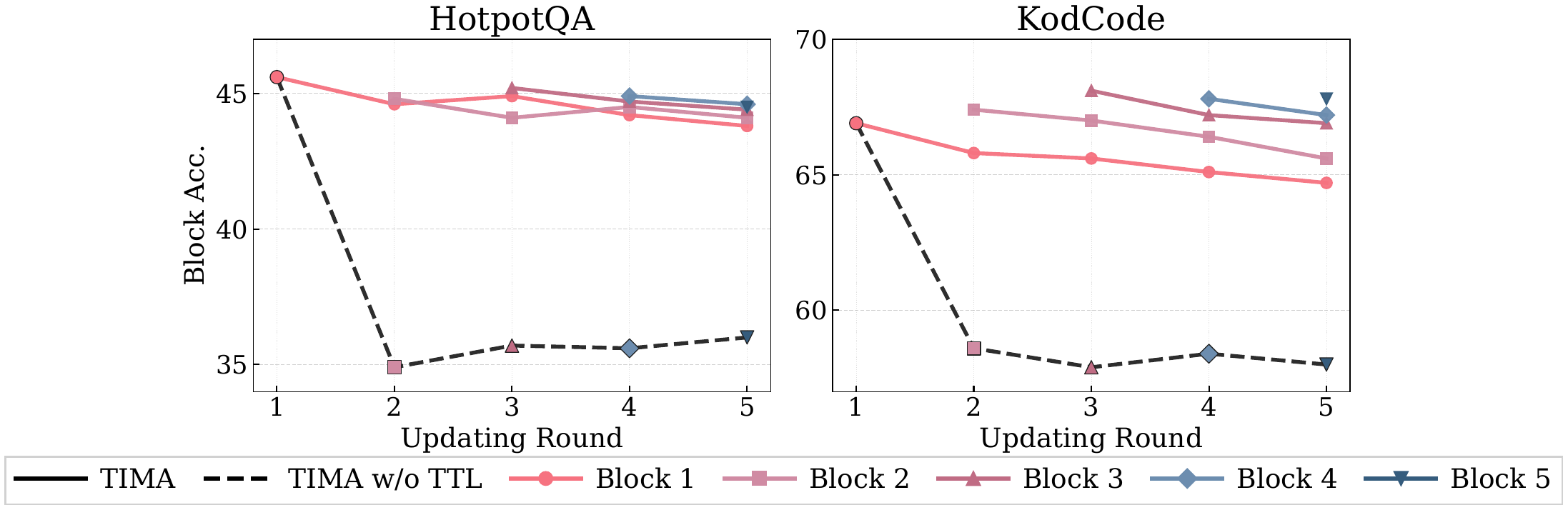}
\caption{Ablation of TTL mechanism under continual updates and frozen \textit{Internalizer} on HotpotQA and KodCode. Each solid curve tracks one block's accuracy across subsequent TTL update rounds; the dashed curve reports the accuracy of an \textit{Internalizer} frozen after one TTL pass on Block~1 and then evaluated on each successive sample block.}
\label{fig:forget_wottl}
\end{figure}

To quantify how much of \sysname{}'s stability against forgetting is attributable to continual TTL updates, we further run a counterfactual in which the \textit{Internalizer} is initialized by one TTL pass over Block~1 and then frozen for the remainder of the stream. Figure~\ref{fig:forget_wottl} plots this frozen variant (dashed) alongside the full \sysname{} (solid). By construction, the two curves coincide at Round~1, which confirms that any subsequent divergence is attributable to the update rule rather than to initialization. Once the \textit{Internalizer} is frozen, accuracy drops sharply as the stream moves to Block~2, and then stabilizes at this reduced level across the remaining blocks. The flatness of the dashed curve after the initial drop is itself informative: it indicates that the loss is not random noise but a systematic mismatch between the frozen memory and the distribution of later blocks, which a frozen \textit{Internalizer} cannot close. In contrast, \sysname{}'s performance is identical to Figure~\ref{fig:forgetting}, only decline mildly over the same horizon, as continual updates re-align the \textit{Internalizer} with the evolving stream. The gap between the two regimes directly measures the contribution of TTL updates beyond retrieval and initialization.

\subsection{More t-sne Visualizations}
\begin{figure}[H]
\centering
\includegraphics[width=0.94\linewidth]{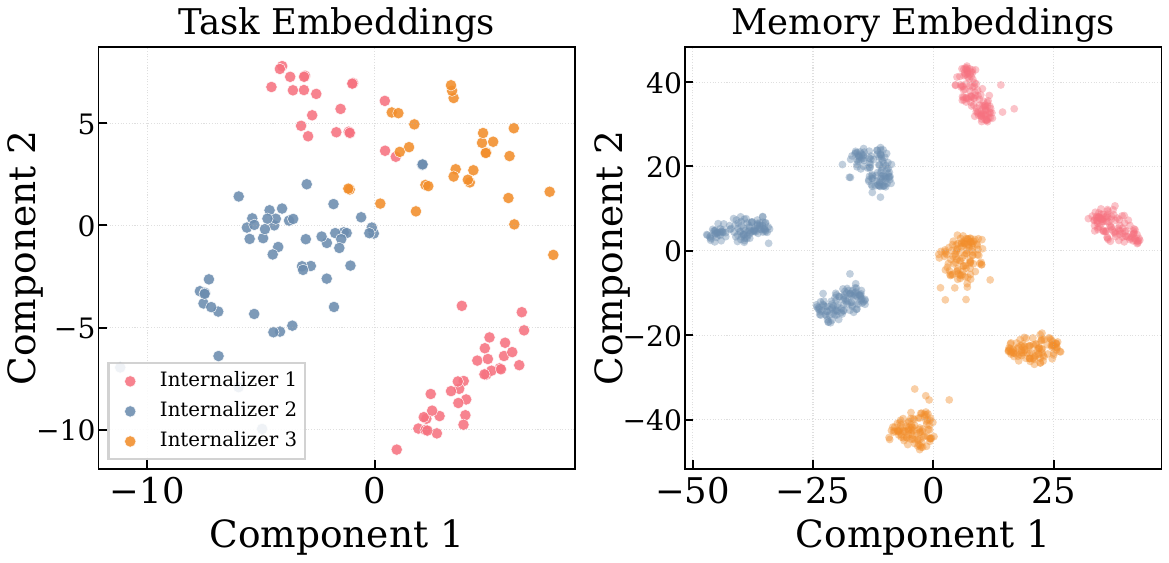}
\caption{t-SNE visualization on KodCode.}
\label{fig:more_tsne}
\end{figure}

\subsection{Efficiency Analysis}
\label{append:efficiency}

\begin{table*}[t]
\centering
\caption{\textbf{Efficiency comparison on Qwen2.5-7B-Instruct.} For \textsc{Tima}, Time and Total TFLOPs are reported as generation + update. Param Effic. denotes the accuracy gain over Vanilla per million trainable parameters.}
\label{tab:efficiency}
\setlength{\tabcolsep}{2.5pt} 
\resizebox{\textwidth}{!}{%
\begin{tabular}{lc ccccc ccccc ccccc}
\toprule
\multirow{2}{*}{\textbf{Method}} & \multirow{2}{*}{\textbf{\makecell{Trainable\\Params}}}
 & \multicolumn{5}{c}{\textbf{KodCode}}
 & \multicolumn{5}{c}{\textbf{GSM8K}}
 & \multicolumn{5}{c}{\textbf{HotpotQA}} \\
\cmidrule(lr){3-7} \cmidrule(lr){8-12} \cmidrule(lr){13-17}
 & & Time (s) & Acc. & \makecell{Param\\Effic.} & \makecell{Output\\Length} & \makecell{Total\\TFLOPs}
   & Time (s) & Acc. & \makecell{Param\\Effic.} & \makecell{Output\\Length} & \makecell{Total\\TFLOPs}
   & Time (s) & Acc. & \makecell{Param\\Effic.} & \makecell{Output\\Length} & \makecell{Total\\TFLOPs} \\
\midrule
\multicolumn{17}{>{\columncolor{modelbg}}l}{\textbf{\textit{Qwen2.5-7B-Instruct}}} \\
Vanilla & 0
 & 14.82 & 52.39 & -- & 325.4 & 8.74
 & 9.48  & 90.33 & -- & 213.2 & 4.97
 & 17.36 & 16.74 & -- & 387.6 & 7.35 \\
MemGen & 30.82M
 & \underline{7.18} & 65.92 & 0.44 & 141.7 & 12.45
 & \underline{7.64} & 91.60 & 0.04 & 135.4 & 9.54
 & \underline{11.29} & 38.56 & 0.71 & 191.0 & 15.93 \\
Titans & 2.17M
 & 10.24 & 53.55 & 0.53 & 206.2 & 5.64
 & 10.72 & 89.30 & $-$0.47 & 219.3 & 5.02
 & 16.42 & 27.07 & 4.76 & 346.9 & 6.38 \\
\rowcolor{timabg}
\textsc{Tima} & 13.21M
 & \makecell{8.46\\[-2pt]{\scriptsize(6.69+1.77)}} & \textbf{66.31} & 1.05 & 146.8 & \makecell{14.43\\[-2pt]{\scriptsize(7.37+7.06)}}
 & \makecell{8.76\\[-2pt]{\scriptsize(6.55+2.21)}} & \textbf{93.47} & 0.24 & 162.2 & \makecell{9.59\\[-2pt]{\scriptsize(5.79+3.80)}}
 & \makecell{13.31\\[-2pt]{\scriptsize(10.58+2.73)}} & \textbf{43.03} & 1.99 & 220.1 & \makecell{10.85\\[-2pt]{\scriptsize(5.76+5.09)}} \\
\bottomrule
\end{tabular}
}
\end{table*}

Table~\ref{tab:efficiency} reports the average inference time and accuracy on KodCode, GSM8K, and HotpotQA. Overall, \sysname{} achieves the best accuracy on all three datasets while maintaining moderate runtime overhead. Compared with MemGen, \sysname{} is slightly slower due to the additional \emph{Internalizer} test-time update, but the overhead remains limited, which is about $15$--$18\%$ across the three datasets. In return, \sysname{} improves accuracy by $+1.87$ and $+4.47$ percentage points against the strongest baseline on GSM8K and HotpotQA, respectively. Compared with Vanilla, \sysname{} is consistently faster, reducing runtime by $42.9\%$ on KodCode and $23.3\%$ on HotpotQA. This suggests that the injected latent memory can guide the backbone toward more focused generation paths, especially for code generation and multi-hop reasoning where unconstrained decoding often leads to longer and less productive trajectories. Titans, despite also performing test-time memory updates, is slower than \sysname{} on all three datasets and yields substantially lower accuracy, indicating that \sysname{} provides a better efficiency--performance trade off.

\subsection{Self-NLL Objective Analysis}
\label{self-nll ana}

\begin{table}[t]
\centering
\caption{\textbf{Correlation between negative entropy and trajectory success.} We report AUROC and Kendall's $\tau_b$ on each dataset.}
\label{tab:entropy_auroc}
\setlength{\tabcolsep}{4pt}
\resizebox{\linewidth}{!}{%
\begin{tabular}{lcccccccc}
\toprule
\textbf{Metric} & \textbf{WebShop} & \textbf{BigCodeBench} & \textbf{GSM8K} & \textbf{MATH} & \textbf{HotpotQA} & \textbf{KodCode} & \textbf{TriviaQA} & \textbf{2WikiMultihopQA} \\
\midrule
\textbf{AUROC}             & 0.7640 & 0.7245 & 0.6794 & 0.6834 & 0.7167 & 0.6633 & 0.7017 & 0.6946 \\
\textbf{Kendall's $\tau_b$} & 0.3411 & 0.3070 & 0.2787 & 0.2940 & 0.3235 & 0.2892 & 0.3024 & 0.2976 \\
\bottomrule
\end{tabular}%
}
\end{table}

Table~\ref{tab:entropy_auroc} examines whether the negative entropy that drives the self-NLL objective is informative about trajectory quality. Across all eight datasets, the negative entropy of a trajectory separates successful from failed trajectories with an AUROC between 0.6633 and 0.7640, well above the chance level of 0.5, and Kendall's $\tau_b$ is positive on every dataset (0.2787--0.3411), so lower entropy is consistently associated with a higher chance of success. The association is strongest on WebShop (AUROC 0.7640, $\tau_b=0.3411$), followed by BigCodeBench and HotpotQA, and weakest on KodCode (0.6633) and GSM8K (0.6794); the two metrics largely agree on this ordering. Entropy is thus an informative but imperfect indicator of correctness, which is why \sysname{} uses it to select the turns used for updates (Eq.~\ref{eq:thresh}) instead of treating every generated turn as supervision.

\begin{table}[t]
\centering
\caption{\textbf{Effect of update-turn selection.} Performance (\%) of \textsc{Tima} when the Internalizer is updated on different subsets of turns. \emph{Accepted turns} uses only the turns that pass the low-entropy filter (default, $E_0 = p_{50}$); \emph{Rejected turns} uses only the turns filtered out; \emph{Updates without filtering} uses all turns. Best results are in bold.}
\label{tab:update_turn_selection}
\resizebox{\linewidth}{!}{%
\begin{tabular}{lcccccccc}
\toprule
\textbf{Update turns} & \textbf{GSM8K} & \textbf{MATH} & \textbf{KodCode} & \textbf{BigCodeBench} & \textbf{2Wiki} & \textbf{HotpotQA} & \textbf{ALFWorld} & \textbf{WebShop} \\
\midrule
Accepted turns            & \textbf{93.47} & \textbf{80.40} & \textbf{66.31} & \textbf{43.27} & \textbf{41.04} & \textbf{43.03} & \textbf{78.1} & \textbf{58.7} \\
Rejected turns            & 83.90 & 57.20 & 48.76 & 33.67 & 34.82 & 28.59 & 48.3 & 32.2 \\
Updates without filtering & 91.57 & 73.40 & 62.51 & 37.88 & 38.08 & 37.10 & 65.7 & 49.7 \\
\bottomrule
\end{tabular}%
}
\end{table}

Table~\ref{tab:update_turn_selection} compares updating the \emph{Internalizer} on different subsets of turns. Updating on the accepted turns, our default, is the best on all eight benchmarks, and updating on the rejected turns is the worst on all eight, with the gap between the two ranging from $6.22$ points on 2Wiki to $29.8$ points on ALFWorld. Since $E_0=p_{50}$ splits the turns into two roughly equal halves, this gap is not caused by a difference in the number of updates but by which turns are used. Updating on all turns without filtering lies in between and trails the accepted-only setting by $1.90$--$12.4$ points, with the largest deficits on ALFWorld ($12.4$) and WebShop ($9.0$), whose long-horizon trajectories contain more turns for noisy updates to accumulate. Updating on rejected turns alone even drops GSM8K and KodCode below the Vanilla results (90.33 and 52.39), whereas updating on accepted turns raises them. These results agree with the AUROC analysis above: low-entropy turns carry a reliable learning signal, while high-entropy turns mostly contribute noise.

\section{Discussion}

\subsection{Limitations}
\label{append:limitation}
Although \sysname{} shows consistent gains across diverse benchmarks, several limitations remain. First, the current framework is evaluated on a finite set of task families and backbone models, and its behavior under substantially broader agent settings, more heterogeneous tool environments, or rapidly shifting task streams remains to be studied. Second, while the \emph{Internalizer} is lightweight, \sysname{} still introduces additional test-time computation due to online memory generation and continual updates, which may become non-negligible in latency-sensitive deployment scenarios. 

Third, the low-entropy selection in \sysname{} acts as a noise filter rather than a correctness filter. Entropy reflects how confident the model is in its own output, not whether that output is correct: the AUROC between negative entropy and trajectory success ranges from 0.66 to 0.76 (Table~\ref{tab:entropy_auroc}), so the filter tends to retain higher-quality turns but cannot exclude confidently wrong ones, which pass the threshold and are reinforced by the self-NLL update. Relatedly, Figure~\ref{fig:forgetting} shows that accuracy still declines as test-time updates accumulate; entropy filtering slows this forgetting compared with updating on all turns, but it does not eliminate it. Incorporating signals beyond the model's own confidence, such as environment feedback or verifiable outcomes, and explicitly constraining the drift of the \emph{Internalizer} over long streams are natural directions for future work. Exploring more efficient update mechanisms, more scalable memory organization, and more robust adaptation is an important direction for future work.

\subsection{Broader Impacts}
\label{append:impacts}
This work studies continually updated latent memory for LLM agents. Such a capability may improve the adaptability, efficiency, and long-horizon consistency of agents in complex environments, with potential benefits for interactive assistants and other agentic systems. At the same time, stronger adaptive memory may also introduce risks, since agents may internalize biased or undesirable patterns from ongoing interaction and may be misused in longer-horizon harmful workflows. We therefore believe that deployment of such systems should be accompanied by appropriate safeguards, including monitoring of memory updates, constraints on update sources, and safety evaluation in realistic settings.

\section{Case Study}
\label{append:case}

\begin{tcolorbox}[
  enhanced,
  breakable,
  sharp corners,
  colback=gray!8,
  colframe=blue!55!black,
  boxrule=1.2pt,
  title={Cases from HotPotQA},
  colbacktitle=blue!55!black,
  coltitle=white,
  fonttitle=\bfseries,
]
{\color{magenta!70!black}\bfseries \small Case 1}\\
{\color{cyan!50!black}\bfseries \small System Prompt}\\
{\ttfamily \small Answer the given multi-hop question. You must reason step by step inside <think> and </think> before every action, and you must think again after each new observation. If you still need evidence, call the search engine with <search> query </search>. The search engine will return documents between <information> and </information>. You may search multiple times. When you have enough evidence, provide only the final short answer inside <answer> and </answer>. For example, <answer> Beijing </answer>.}\\[3pt]
{\color{green!60!black}\bfseries \small User Prompt}\\
{\ttfamily \small Aside from the Apple Remote, what other device can control the program Apple Remote was originally designed to interact with?}\\[3pt]
{\color{blue!70!black}\bfseries \small Assistant Output}\\
{\color{orange!90!black}\ttfamily \bfseries \small [TIMA Tokens]}{\ttfamily \small <think> To answer this question, I need to know what program the Apple Remote was originally designed for and if there are other devices that can control it.</think> <search> original program for Apple Remote </search>}\\[3pt]
{\color{green!60!black}\bfseries \small Information}\\
{\ttfamily \small Doc 1(Title: Siri Remote) The Siri Remote is a remote control device released in tandem with the 4th Generation Apple TV at an Apple Keynote Address by CEO Tim Cook on October 26, 2015 by Apple Inc. The Remote is designed specifically to work with the always-on Siri capable Apple TV...[truncted]}\\[3pt]
{\color{blue!70!black}\bfseries \small Assistant Output}\\
{\color{orange!90!black}\ttfamily \bfseries \small [TIMA Tokens]}{\ttfamily \small <think> From the information provided, the Apple Remote was originally designed to interact with the Front Row media program on the iSight iMac G5 and is compatible with some later desktop and portable Macintosh computers. </think> <search> devices compatible with Front Row media program </search>}\\[3pt]
{\color{green!60!black}\bfseries \small Information}\\
{\ttfamily \small Doc 1(Title: Front Row (software)) Front Row is a discontinued media center software application for Apple's Macintosh computers and Apple TV for navigating and viewing video, photos, podcasts, and music from a computer, optical disc, or the Internet through a 10-foot user interface...[truncted]}\\[3pt]
{\color{blue!70!black}\bfseries \small Assistant Output}\\
{\color{orange!90!black}\ttfamily \bfseries \small [TIMA Tokens]}{\ttfamily \small <think> The information confirms that the Apple Remote was designed to interact with the Front Row media program. The Front Row media program can be controlled by an Apple Remote or the keyboard function keys. </think> <answer> keyboard function keys </answer>}\\[3pt]
{\color{magenta!70!black}\bfseries \small Case 2}\\
{\color{cyan!50!black}\bfseries \small System Prompt}\\
{\ttfamily \small Answer the given multi-hop question. You must reason step by step inside <think> and </think> before every action, and you must think again after each new observation. If you still need evidence, call the search engine with <search> query </search>. The search engine will return documents between <information> and </information>. You may search multiple times. When you have enough evidence, provide only the final short answer inside <answer> and </answer>. For example, <answer> Beijing </answer>.}\\[3pt]
{\color{green!60!black}\bfseries \small User Prompt}\\
{\ttfamily \small What material did a hairdresser from Yorkshire, England invent that was named by his granddaughter?}\\[3pt]
{\color{blue!70!black}\bfseries \small Assistant Output}\\
{\color{orange!90!black}\ttfamily \bfseries \small [TIMA Tokens]}{\ttfamily \small <think> To answer this question, I need to identify the hairdresser from Yorkshire, England, and the material he invented that was named by his granddaughter. I will start by searching for a notable hairdresser from Yorkshire, England. </think> <search> notable hairdressers from Yorkshire, England </search>}\\[3pt]
{\color{green!60!black}\bfseries \small Information}\\
{\ttfamily \small Doc 1(Title: Samuel Ironside) Samuel Ironside (9 September 1814–24 April 1897) was an English Methodist missionary in New Zealand, where he became a supporter and signatory of the Treaty of Waitangi...[truncted]}\\[3pt]
{\color{blue!70!black}\bfseries \small Assistant Output}\\
{\color{orange!90!black}\ttfamily \bfseries \small [TIMA Tokens]}{\ttfamily \small <think> The search results do not mention a hairdresser from Yorkshire, England who invented a material named by his granddaughter. I need to narrow down the search to hairdressers specifically and look for any inventions or materials they might have created. </think> <search> hairdressers from Yorkshire, England inventions </search>}\\[3pt]
{\color{green!60!black}\bfseries \small Information}\\
{\ttfamily \small Doc 1(Title: List of English inventions and discoveries) English inventions and discoveries are objects, processes or techniques invented...[truncted]}\\[3pt]
{\color{blue!70!black}\bfseries \small Assistant Output}\\
{\color{orange!90!black}\ttfamily \bfseries \small [TIMA Tokens]}{\ttfamily \small <think> The search results still do not provide the specific information I need. I need to look for a hairdresser from Yorkshire, England who invented a material and had a granddaughter involved in naming it. I will try a more specific search. </think> <search> Yorkshire hairdresser invented material named by granddaughter </search>}\\[3pt]
{\color{green!60!black}\bfseries \small Information}\\
{\ttfamily \small Doc 1(Title: Starlite) Starlite is a material claimed to be able to withstand and insulate from extreme heat. It was invented by amateur chemist and hairdresser Maurice Ward (1933–2011) during the 1970s and 1980s...[truncted]}\\[3pt]
{\color{blue!70!black}\bfseries \small Assistant Output}\\
{\color{orange!90!black}\ttfamily \bfseries \small [TIMA Tokens]}{\ttfamily \small <think> The search results provide the information I need. Maurice Ward, an amateur chemist and hairdresser from Yorkshire, England, invented Starlite, a material claimed to withstand and insulate from extreme heat. The name Starlite was coined by his granddaughter, Kimberly. </think> <answer> Starlite </answer>}\\[3pt]

\end{tcolorbox}



\end{document}